\pdfoutput=1

\documentclass[11pt,a4paper]{article}

\usepackage[preprint]{acl}

\usepackage{times}
\usepackage{latexsym}

\usepackage[T1]{fontenc}

\usepackage[utf8]{inputenc}

\usepackage{microtype}

\usepackage{inconsolata}

\usepackage{graphicx}
\usepackage{subfig}   
\usepackage[inline]{enumitem}

\newcommand{\npapersreviewed}{$64$}

\definecolor{concept}{RGB}{239,205,138}
\definecolor{philosophy}{RGB}{170,190,234}
\definecolor{manifestation}{RGB}{239,181,188}
\definecolor{implementation}{RGB}{208,227,175}

\usepackage{booktabs}
\usepackage[most]{tcolorbox}
\usepackage{xcolor}
\usepackage{csvsimple}
\usepackage{pifont}

\definecolor{pastelyellow}{HTML}{F9EED6}
\definecolor{darkyellow}{HTML}{B07537}

\definecolor{pastelblue}{HTML}{DBE9FB}
\definecolor{darkkblue}{HTML}{5D81B8}

\definecolor{pastelred}{HTML}{F8ECEB}
\definecolor{darkred}{HTML}{9E6780}

\definecolor{pastelpurple}{HTML}{F2EBF8}
\definecolor{darkpurple}{HTML}{7E6D96}

\definecolor{pastelgreen}{HTML}{E7ECE3}
\definecolor{darkgreen}{HTML}{6B8836}

\newlist{gitemize1}{itemize}{4}
\setlist[gitemize1,1]{
  leftmargin=\dimexpr0.5cm+\labelsep\relax,
  label={\smash{\raisebox{-0.25\height}{\includegraphics[width=0.5cm]{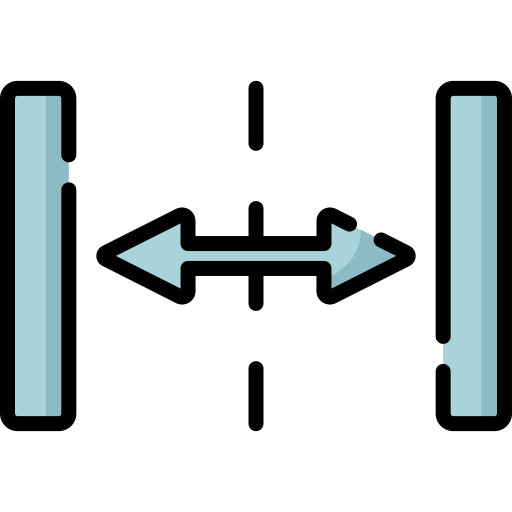}}}}
}

\newlist{gitemize2}{itemize}{4}
\setlist[gitemize2,1]{
  leftmargin=\dimexpr0.3cm+\labelsep\relax,
  label={\smash{\raisebox{-0.25\height}{\includegraphics[width=0.5cm]{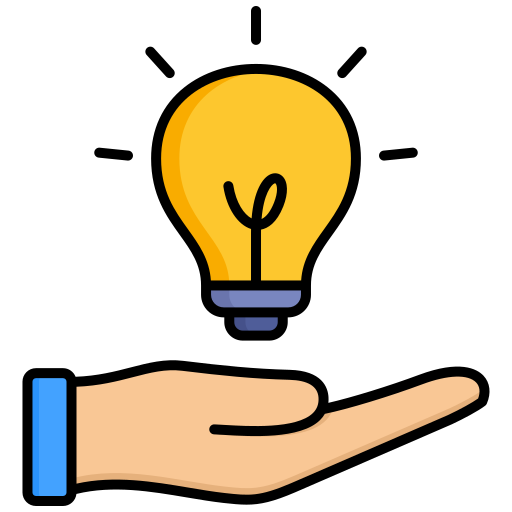}}}}
}

\usepackage{xspace}

\title{Not All Subjectivity Is the Same! \\ Defining Desiderata for the Evaluation of Subjectivity in NLP}

\author{
 \textbf{Urja Khurana$^{\includegraphics[width=0.02\textwidth]{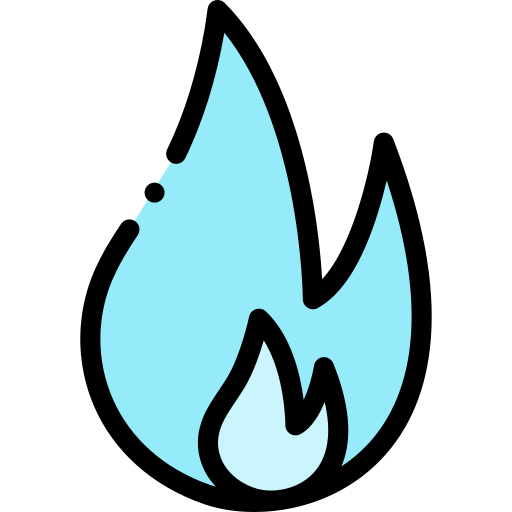}}$},
 \textbf{Michiel van der Meer$^{\includegraphics[width=0.02\textwidth]{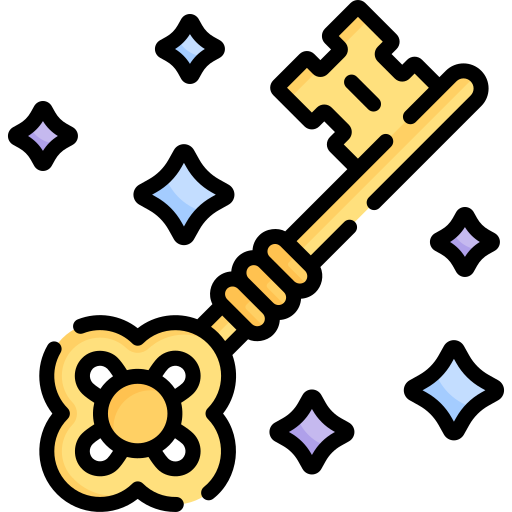}}$}\thanks{Equal contribution.},
\textbf{Enrico Liscio$^{\includegraphics[width=0.02\textwidth]{figures/emojis/natural-gas.png}}$}$^*$, \\
 \textbf{Antske Fokkens$^{\includegraphics[width=0.02\textwidth]{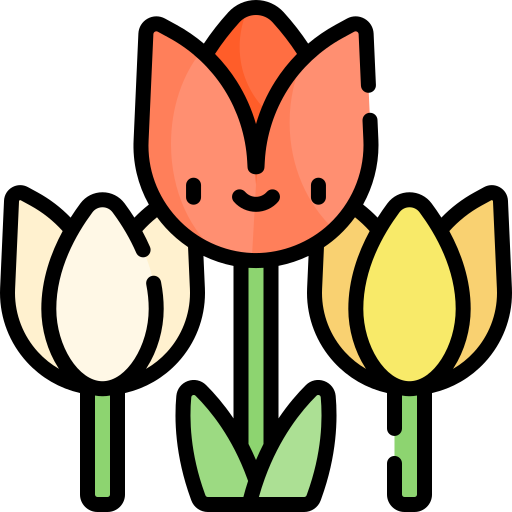}}$}
  \textbf{Pradeep K. Murukannaiah$^{\includegraphics[width=0.02\textwidth]{figures/emojis/natural-gas.png}}$}
\\
  {\includegraphics[width=0.02\textwidth]{figures/emojis/natural-gas.png}}Department of Intelligent Systems, Delft University of Technology\\
{\includegraphics[width=0.02\textwidth]{figures/emojis/key.png}}Leiden Institute of Advanced Computer Science (LIACS), Leiden University\\
{\includegraphics[width=0.02\textwidth]{figures/emojis/tulip.png}}Computational Linguistics and Text Mining Lab, Vrije Universiteit Amsterdam\\
\texttt{\{u.khurana, e.liscio, p.k.murukannaiah\}@tudelft.nl,} \\ \texttt{m.t.van.der.meer@liacs.leidenuniv.nl, antske.fokkens@vu.nl}
}

\begin{document}
\maketitle

\begin{abstract}
    Subjective judgments are part of several NLP datasets and recent work is increasingly prioritizing models whose outputs reflect a diversity of perspectives. Such responses allow us to shed light on minority voices, which are frequently marginalized or obscured by dominant perspectives. It remains a question whether our evaluation practices align with these models’ objectives. This position paper proposes \textit{seven} evaluation desiderata for subjectivity-sensitive models, rooted in how subjectivity is represented in NLP data and models. The desiderata are constructed in a top-down approach, keeping in mind the user-centric impact of such models. We scan the experimental setup of \npapersreviewed~papers and show that various aspects of subjectivity are still understudied: the distinction between ambiguous and polyphonic input, whether subjectivity is \textit{effectively} expressed to the user, and a lack of interplay between different desiderata, amongst other gaps. 
\end{abstract}

\section{Introduction}
For several NLP tasks, such as answering general knowledge questions, there is a \textit{factual} or \textit{correct} response that is generally accepted. However, for some tasks, the ``truth'' depends on the context or individual interpretation. Consider, for example, the task of determining whether an online comment is offensive. There is no universally agreed notion of \textit{offensiveness} or \textit{hate speech} \citep{vidgen-etal-2019-challenges, fortuna-etal-2020-toxic, leonardelli-etal-2021-agreeing, khurana-etal-2022-hate}. Instead, the sociocultural background and experiences of the readers influence their judgments \citep{sap-etal-2022-annotators, toxicity-goyal-2022}. Similarly, judgments depend on a personal interpretation in many other tasks such as natural language inference \citep{pavlick-kwiatkowski-2019-inherent, nie-etal-2020-learn}, semantic annotation \citep{sommerauer-etal-2020-describe}, and understanding indirect answers \citep{damgaard-etal-2021-ill}. We refer to such tasks that have multiple plausible labels rather than a single valid ground-truth as \textit{\textbf{subjective}}. 

\begin{figure*}[h!]
    \centering
    \includegraphics[width=\linewidth]{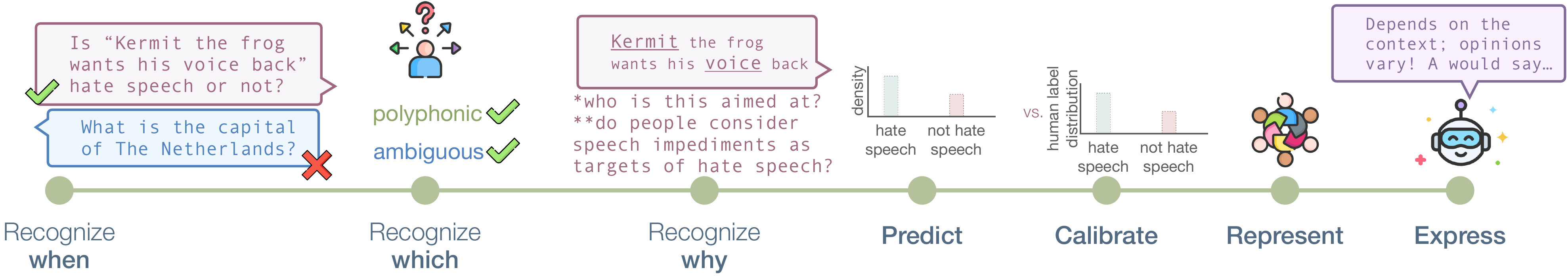}
    \caption{The sequential nature of our proposed desiderata for evaluating a subjectivity-sensitive model.}
    \label{fig:overview}
\end{figure*}

As language models are integrated into societal applications, they increasingly engage with subjective concepts such as values, norms, and arguments \citep{forbes-etal-2020-social, lourie2021scruples, liscio-etal-2023-text, van-der-meer-etal-2024-empirical}. Therefore, their behavior must align with varying human preferences \citep{sorensen2024position} and reflect multiple viewpoints \citep{cabitza2023toward, plank-2022-problem}. Yet, such models have been shown to collapse to a specific view \citep{hayati-etal-2024-far}, typically as a consequence of how labels are presented (aggregated to a single majority label) or due to prevalent preference-tuning paradigms (such as RLHF \citep{rlhfpaper} and DPO \citep{dpopaper}) that optimize for the majority vote or general preference \citep{casper2023open}. 
Obfuscating minority or diverging viewpoints is problematic when deploying models in real-world settings, running the risk of further disenfranchising marginalized communities \citep{xu-etal-2021-detoxifying}.

\textit{Subjectivity-sensitive models} reflect variation in judgment and enable the representation of minority perspectives. 
As the NLP field pushes for such models \citep{ovesdotter-alm-2011-subjective, basile-etal-2021-need, plank-2022-problem, cabitza2023toward, wang-etal-2025-perspective}, a critical look at their evaluation practices is necessary.
Several aspects affect the evaluation of subjectivity-sensitive models, including the ideals we outline, the extent to which subjective evaluation metrics capture overall trends or fine-grained model behavior \citep{chen-etal-2024-seeing, rizzi-etal-2024-soft}, and the type of task (classification vs. generation). To this end, we address the following question:
\textbf{which aspects of subjectivity should evaluation metrics capture?}

To address this question, in this position paper, we first review how subjectivity is represented in the data and incorporated into NLP models, highlighting a previously underexplored aspect: the distinction between ambiguity (input that can warrant multiple interpretations) and polyphony (input that has multiple valid perspectives). These insights further drive our top-down approach, in which we propose a set of sequential desiderata to evaluate subjectivity-sensitive models (outlined in Figure~\ref{fig:overview}). We identify aspects not yet addressed in the literature by drawing on insights from related fields. Finally, we map these desiderata onto current evaluation practices, revealing gaps and opportunities for future work.

Our key contributions are threefold: (1) desiderata for evaluating subjectivity-sensitive models, (2) a systematic and indicative categorization of current evaluation metrics, and (3) identified gaps and opportunities. 

\section{Subjectivity in Data}
\label{sec:subjectivity-in-data}
 
To establish a fitting evaluation paradigm, we start by describing how subjectivity and related concepts are interpreted in the NLP field. Next, we discuss what the \emph{sources} of subjectivity are. Finally, we reflect on the types of information to incorporate in a subjectivity-sensitive dataset.

\subsection{Subjectivity and Related Concepts}
\label{subsec:defining-subjectivity}
Subjectivity pertains to the existence of multiple valid ground-truths \citep{vardomskaya2018sources, rottger-etal-2022-two, plank-2022-problem}. NLP literature also discusses other concepts related to subjectivity. \textit{Perspectivism} \citep{cabitza2023toward, frenda2025perspectivist} recognizes that ground-truth depends on individual perspectives. This has taken shape as a \textit{perspectivist} framework for representing different perspectives in data: 
\textit{weak} when multiple perspectives are considered only during annotation, and \textit{strong} when extending this to incorporating multiple perspectives in the model output as well. Subjective interpretations can lead to \textit{disagreement} \citep{uma2021learning} when different individuals have opposing views on a specific input. \citet{plank-2022-problem} uses the term \textit{human label variation} to refer to instances with different plausible perspectives, since \textit{disagreement} could imply that at least one judgment is incorrect. \textit{Disagreement} can also arise in factual matters, capturing annotator \textit{errors} due to misunderstanding (misinformation or misinterpreting annotation guidelines) or attention drifts. 

\subsection{Sources of Subjectivity}
\label{subsec:sources-subjectivity}
Where do multiple subjective viewpoints or beliefs stem from? \citet{aroyo2015truth} suggest three potential sources of disagreement in their Triangle of Reference framework that maps out the process of interpreting an input: \textit{task ambiguity}, \textit{input ambiguity}, and \textit{annotator attitude} \citep{jiang-marneffe-2022-investigating,liu-etal-2023-afraid}. For \textit{task ambiguity}, imprecise task definitions and annotation guidelines may (un)intentionally leave enough space for an annotator to fill the gaps through their own interpretation \citep{rottger-etal-2022-two}. For instance, in ``\textit{\underline{Berlin} decided ...}'', the entity \textit{Berlin} can be annotated as a location or an organization. Without explicitly guiding annotators in such scenarios (in this case, where a city name is used to refer to a government), unintended disagreement can arise.  
Even when the task is clearly defined, \textit{input ambiguity} may still arise when more context is needed for an accurate judgment. Consider an input that contains a reclaimed slur; whether it is hate speech requires the knowledge of who is behind the piece of text. Finally, \textit{annotator-specific factors} such as an annotator's demographic, emotional state, and experience can yield different valid judgments \cite{sap-etal-2022-annotators}, particularly when there is no universally agreed-upon ground-truth.

\subsection{Incorporating Subjectivity in Data}
\label{subsec:incorporating-subj-data}
To train and evaluate models that can handle subjectivity, adequate datasets should be available. 
Conventionally, NLP datasets are built by collecting $n$ annotations per sample and aggregating them into a single ``gold'' label, e.g., through majority voting. However, when there is variation in annotator response, this results in a loss of information on human disagreement. This has led to calls for the release of raw annotations \citep{prabhakaran-etal-2021-releasing, cabitza2023toward, pandita2025forest}, with recent work showing the importance of having a large number of annotations per sample \citep{khurana2024crowdcalibrator, gruber-etal-2024-labels}.

To model the opinion of an individual (annotator), we would need demographic information \citep{bender-friedman-2018-data, prabhakaran-etal-2021-releasing} and a sufficient amount of annotated samples. Disagreement can also stem from annotation mistakes (e.g., due to attention drift), requiring methods \citep{disagreement-deconvolution-gordon-2021, weber-genzel-etal-2024-varierr, jinadu-ding-2024-noise, ivey-etal-2025-nutmeg} and metrics \citep{dumitrache2018crowdtruth, abercrombie-etal-2025-consistency} that can distinguish such \textit{noise} from actual signal (i.e. ways to distinguish variation due to errors from variation reflecting differences in perspective) and indicate annotator reliability. 

The part of the subjective spectrum that is captured is influenced by the pool of annotators.
Smarter annotation techniques better capture how an input is perceived, e.g., by having a variety of demographics annotate a specific sample \citep{pei-jurgens-2023-annotator,van-der-meer-etal-2024-annotator}, selecting annotators of a specific target group \citep{toxicity-goyal-2022},
or a framework for subjectivity throughout the annotation stages \citep{crowdworksheets-diaz-2022}.

\section{Dimensions of Subjective Input}
\label{sec:dimensions}
\begin{figure}[t]
    \centering
    \includegraphics[width=\linewidth]{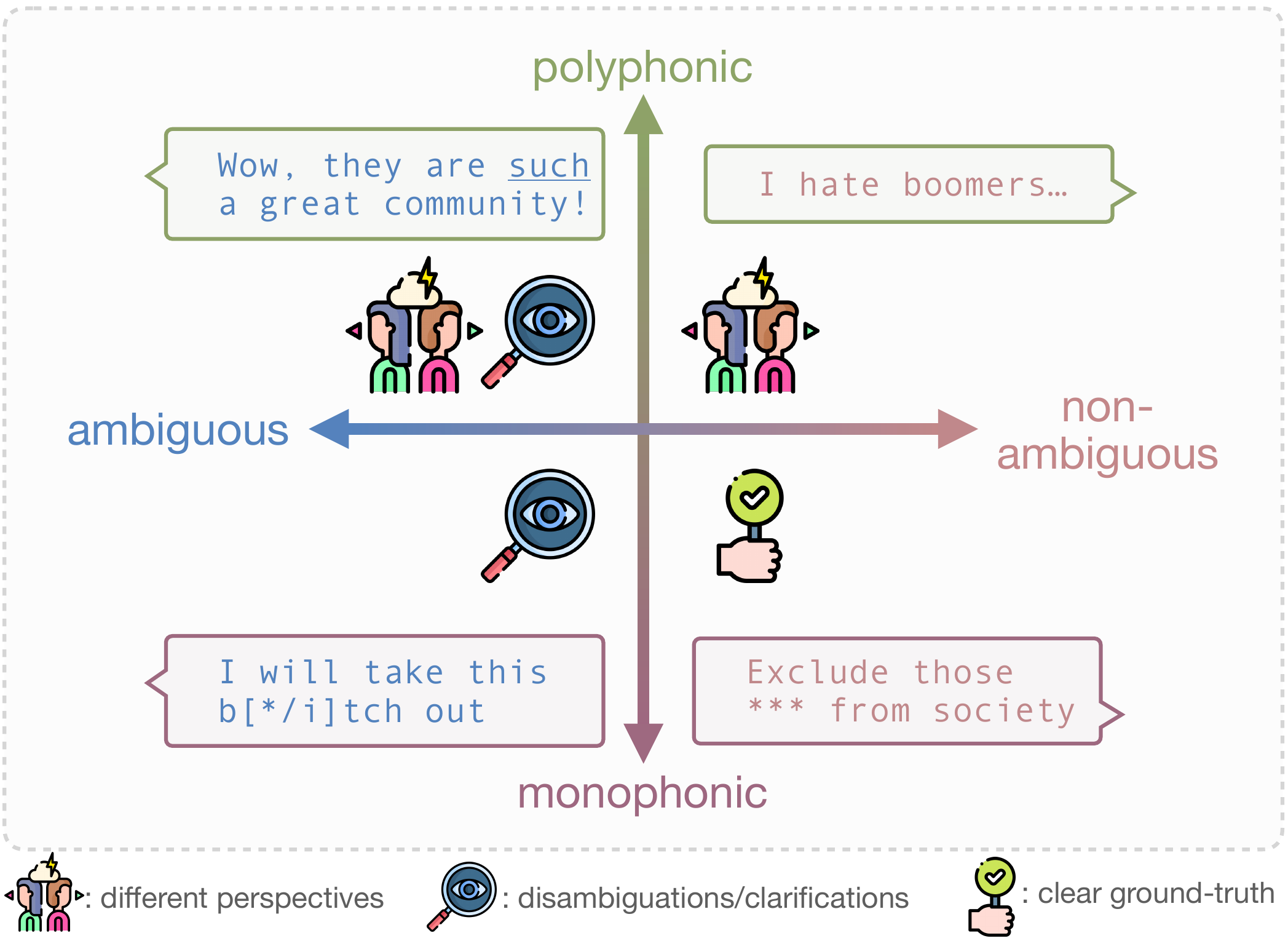}
    \caption{Dimensions of sources of subjectivity at a \textbf{sample-level}: \textit{ambiguity} (x-axis) and \textit{polyphony} (y-axis). Each quadrant shows the desired model output (as icons) with examples situated in hate speech detection.}
    \label{fig:subjectivity-axes}
\end{figure}

What do the sources of subjectivity mean for a subjectivity-sensitive model? We take inspiration from the Triangle of Reference framework \citep{aroyo2015truth} discussed earlier. Where \textit{task ambiguity} plays an overarching role when collecting ground-truth labels across multiple samples, \textit{input ambiguity} and \textit{annotator attitude} can primarily occur at a sample level. These two sources require the model to handle subjectivity differently, an aspect that has not yet been addressed in previous research. We characterize model behavior when dealing with different sources of subjectivity on two dimensions: \textbf{ambiguity} and \textbf{polyphony} (Figure~\ref{fig:subjectivity-axes}).

When an input is non-ambiguous and monophonic, a model can directly incorporate the objective ground-truth. However, modeling scenarios involving subjective factors is more challenging as the following subsections illustrate.

\subsection{Ambiguous $\leftrightarrow$ Non-Ambiguous} 
An input is \textit{ambiguous} when it can warrant different interpretations because the information (e.g., context) needed to make a judgment is incomplete. Inputs that can be interpreted without filling information gaps are \textit{non-ambiguous}. 

\paragraph{Ambiguous.} When faced with an ambiguous input, a model should identify the spectrum of different possible interpretations and disambiguations. For instance, in \textit{``I will take this b\{*/i\}tch out}'' the ambiguity lies in who the \textit{b-word} refers to; the slur or the dog. 

\paragraph{Non-ambiguous.} A non-ambiguous input removes the requirement for a model to understand the different interpretations. In such a situation, the model can directly give a suitable response based on the exact information outlined in the input.

\subsection{Polyphonic $\leftrightarrow$ Monophonic} 
Independent of ambiguity, certain inputs may not have an agreed-upon ground truth, allowing for judgments based on an individual's interpretation. Personal taste, demographics, background, experience, and other factors influence this interpretation. We consider such cases to be \textit{polyphonic}: different perspectives co-existing. Where there is unanimity, we consider that to be \textit{monophonic}.

\paragraph{Polyphonic.} A polyphonic input is open to multiple valid perspectives, which the model should capture. Take for example the statement \textit{``I hate boomers''}; the meaning is clear but the polyphony lies in whether boomers are considered targets of hate speech. A model that can capture polyphony for the right reasons will be able to generalize well in the real world. Thus, the model should recognize \textit{what} aspect(s) of the input enable different perspectives and \textit{why}, e.g., is it the gender or culture that potentially influences the opinion or are there other factors? Not every different perspective has to be attributed to specific demographics, as deviations can also be due to individual differences.  

\paragraph{Monophonic.} When an input has a single valid belief, often connected to factuality, the model can directly predict the ground-truth without the need to consider a variety of potential perspectives. For instance, the phrase \textit{``Exclude those *** from society''} should not warrant any subjective interpretation due to its overt form of hate speech.

\subsection{Ambiguous and Polyphonic}
Ambiguity and polyphony can be intertwined. Consider an input that lacks context \underline{\textit{and}} warrants multiple perspectives, e.g., \textit{``Wow, they are such a great community''}. Here, the ambiguity lies in the sentiment of the sentence and the polyphony in the gravity of the (potential) negative sentiment. In such a scenario, a model is expected to capture the appropriate disambiguation and clarification questions, and the different possible perspectives. 

\section{Subjectivity in Models}
\label{sec:subjectivity-in-models}

How a model expresses subjectivity influences its evaluation. Thus, we briefly discuss ways of expressing and instilling subjectivity in models.
Most papers on subjectivity focus on modeling, which we refer to for detailed discussions \citep{frenda2025perspectivist, xu2026beyond}. We notice a clear lack of work surrounding how to evaluate subjectivity, which we expound on in Sections~\ref{sec:desiderata} and \ref{sec:current-eval-practices}. 

\subsection{Expressing Subjectivity}
\label{subsec:expressing-subjectivity}

To move away from rigid responses in line with the majority preference, the model's output should reflect the spectrum of different perspectives or be aligned to an individual's perspective; \textit{pluralistically aligned}. \citet{sorensen2024position} operationalize this as follows:
(1) \textbf{Distributional}: the model outputs the distribution of perspectives. This can be expressed in the logits or softmax distribution of the model, or directly in the generated response \citep{meister-etal-2025-benchmarking}. 
(2) \textbf{Overton}: the output summarizes different \textit{reasonable} perspectives.
The summary should accurately reflect the window of perspectives.
(3) \textbf{Steering}: the model considers multiple perspectives and values and steers its response to the user's. This is closely related to \textit{personalization}, where preferences go beyond values and include matters such as \textit{textual} or \textit{linguistic} style \citep{zhang2025personalization}.

\subsection{Instilling Subjectivity}
\label{subsec:instilling-subjectivity}
We categorize approaches based on whether subjectivity is instilled when training or fine-tuning, during specific alignment steps, or during inference. At \textbf{inference-time}, integrating subjectivity is usually more computationally efficient than at the training/fine-tuning stage, particularly in the case of LLMs. A popular approach is persona prompting \citep{santurkar2023whose, deshpande-etal-2023-toxicity}, where the model is instructed in a prompt to impersonate a specific \textit{persona}. This can span different demographics where the persona could steer toward a specific perspective or provide a spectrum of multiple perspectives. 
Typically, for subjective modeling during \textbf{fine-tuning} we use \textit{soft} labels, i.e., the distribution of class annotations for a single sample \citep{jamison-gurevych-2015-noise, uma-etal-2021-semeval}. Instead of optimizing for a single (aggregated) \textit{hard} label, we minimize the distance between the estimated and the actual distribution. For this we use distance-based loss functions such as \textit{cross-entropy}, \textit{Kullback-Leibler divergence}, or \textit{Jensen-Shannon divergence}. 
We can also learn from individual annotators' labels to make sets of predictions that constitute annotation distributions \citep{rodrigues2018deep}. 
\textbf{Preference tuning} methods for large language models primarily model majority preference \citep{casper2023open} but strategies are being developed to reflect diverse preferences \citep{zhao2024group, poddar2024personalizing}. 

\section{Desiderata for Subjective Models}
\label{sec:desiderata}
After outlining the desired model output when facing a subjective input, the natural next step is to evaluate its performance. \textit{But what does it mean for a model to perform well on subjective input?} 

To answer this question, we propose desiderata for a subjectivity-sensitive model. These desiderata are motivated by three overarching ideals rooted in our preceding discussion of how subjectivity is represented in NLP data and models: (1) representation of all, including minority voices, (2) based on the correct sources of subjectivity in the input, and (3) the representation of subjectivity (in data and models) generalizing to the real world. 

\paragraph{\texttt{D1.}\texttt{Recognize when subjective modeling is necessary.}}

    \noindent As we highlight in Section \ref{subsec:defining-subjectivity}, inputs may elicit multiple valid perspectives. For a model to make predictions that reflect this diversity, the first step is to \textbf{recognize when} an input needs to be treated subjectively versus objectively. Even when a task is generally considered to be subjective, this step is vital---there will be cases where raters will have consensus, e.g., when dealing with a severe case of hate speech; predicting an aggregate estimate would be sufficient in that case. 

\paragraph{\texttt{D2.}\texttt{Recognize which type of subjective modeling is needed.}}

    \noindent As we describe in Section~\ref{sec:dimensions}, model behavior depends on the type of subjective input. Thus, a model should \textbf{recognize which} type of subjectivity the input contains: ambiguity, polyphony, or both. 

\paragraph{\texttt{D3.}\texttt{Recognize why there is subjectivity in the input.}}

    \noindent Building on the desired model behavior we outline in Section~\ref{sec:dimensions}, a model should \textbf{recognize why} there is ambiguous or polyphonic subjectivity in the input. Using this information, the model can delineate the different disambiguations, clarifications, or perspectives to resolve. This facilitates better predictions and more reliable generalization to unseen scenarios.  

\paragraph{\texttt{D4.}\texttt{Predict subjectivity accurately.}}

\noindent After recognizing the source of the subjectivity, a model should predict the subjectivity correctly.
A model can \textbf{predict} subjectivity in different ways, as we discussed in Section~\ref{subsec:expressing-subjectivity}.

\paragraph{\texttt{D5.}\texttt{Calibrate the output with respect to the frequency of opinions.}}

\noindent As illustrated in Section~\ref{subsec:instilling-subjectivity}, certain models are trained to estimate the distribution of human judgment for an input. Depending on how subjectivity is expressed (distributional or overton), the predicted distribution should be \textbf{calibrated} to the actual human judgment distribution. This applies both when a probability distribution is obtained directly from the model and when an overton response (where linguistic quantifiers should be well-calibrated) is generated. 

\paragraph{\texttt{D6.}\texttt{Represent all perspectives.}}

\noindent The model should account for the minority voices that are usually lost in aggregation, by \textbf{representing} diverse viewpoints. 
Relevant perspectives should be decoupled from dominant perspectives. For example, it should not only echo a Western perspective but also other cultures when the input warrants it.

\paragraph{\texttt{D7.}\texttt{Express subjectivity without hampering user experience.}}

\noindent A subjectivity-sensitive model carries with it the responsibility of \textit{informing} a user of different perspectives. This mirrors the responsibility of showing users explanations of decision-making in the field of explainable AI (XAI). What counts as a ``good'' explanation depends on the user \citep{ehsan-2020-hcxai}.
In a similar vein, we argue that subjectivity should be \textbf{expressed} in a way that does not hamper user experience. 
A response that enumerates all potential perspectives can overload the user.\footnote{This excludes a \textit{distributional} output from the model.} Though overload can sometimes be prevented by structuring responses, this desideratum can also be at odds with \textbf{\texttt{D6}}. Ideally, the perspectives included in a response should be \textit{useful}. However, this is challenging as what is useful is user- and context-dependent (e.g., to what extent do we consider minority opinions, even when \textit{one} person holds it? Does a user find it important to expand their worldview? Or would they rather know more about perspectives from similar cultures? And is what a user wants always that what the user needs?). To summarize, usefulness is a complex notion and what this means in practice requires more research.

\begin{figure}[h!]
    \centering
    \includegraphics[width=\linewidth]{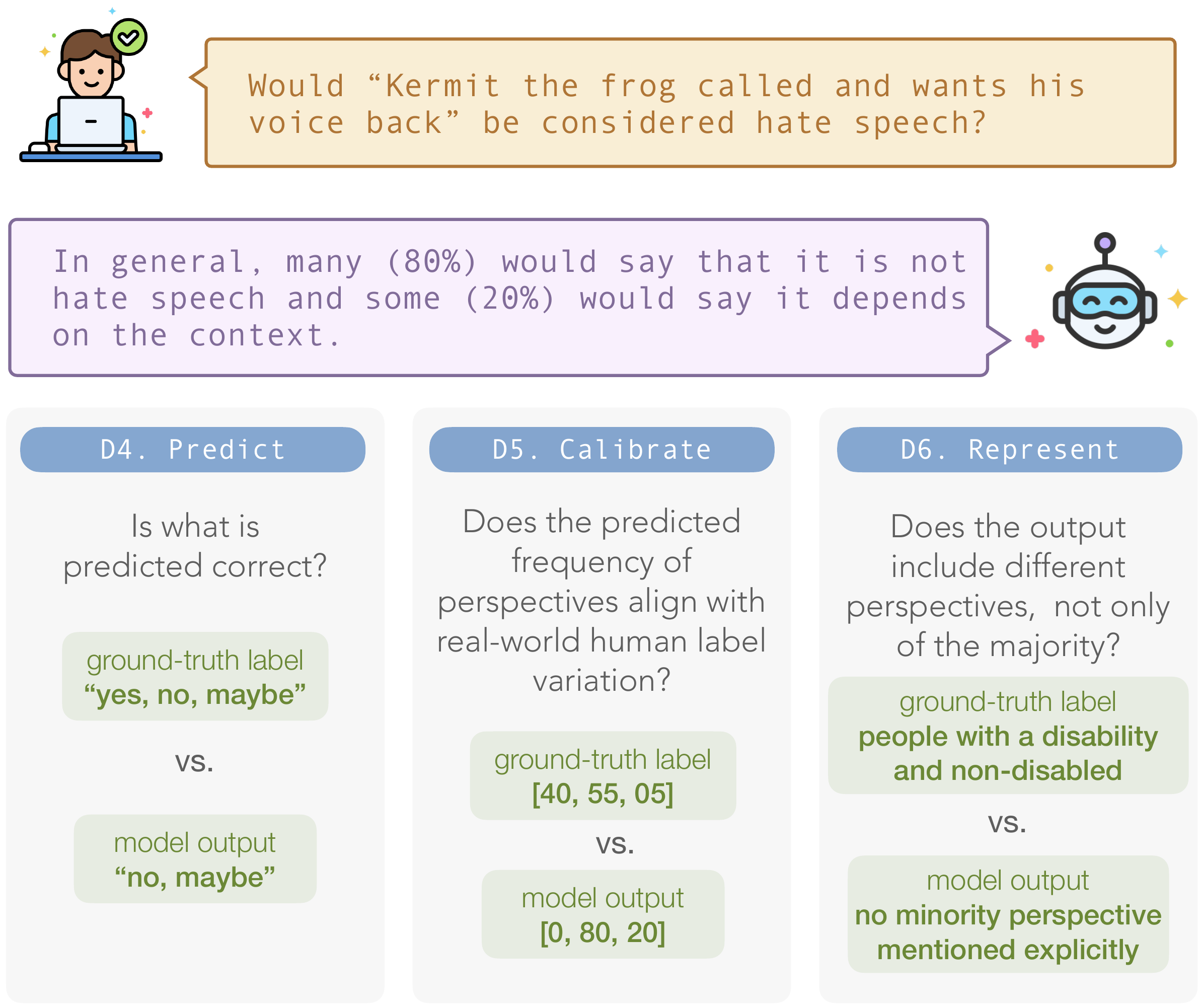}
    \caption{Complementary nature of the desiderata that specifically evaluate the model output from different angles: \textbf{\texttt{D4. Predict}}, \textbf{\texttt{D5. Calibrate}}, and \textbf{\texttt{D6. Represent}}. The example (including all the values for the labels) is for illustrative purposes only, showcasing what each desideratum captures. In this case the input text (taken from \citet{schmidt-wiegand-2017-survey}) is highly contextual and could potentially be harmful to people with a speech disability. Each desideratum focuses on a different angle, though \textbf{\texttt{D4}} and \textbf{\texttt{D5}} can have some overlap if distance metrics are used for \textbf{\texttt{D4}}.}
    \label{fig:desiderata-boundaries}
\end{figure}

\paragraph{Practical considerations.} Our desiderata can be evaluated sequentially and sometimes also independently. They allow for a mix-and-match approach depending on the nature and setup of the task at hand. Certain desiderata are connected by looking at the model output from different angles. We illustrate this complementary nature of desiderata \textbf{\texttt{D4}}, \textbf{\texttt{D5}}, and \textbf{\texttt{D6}} in Figure~\ref{fig:desiderata-boundaries}. Note that the model does not have to be explicitly designed to address all of the desiderata, but can capture them implicitly as well (e.g., a model does not have to be explicitly designed to output binary labels for \textbf{\texttt{D1}}, it can cover this desideratum implicitly when directly predicting the human judgment distribution). The application also influences to what extent all the desiderata should be covered by the model, as well as the order of certain desiderata. For instance, for an NLI model doing requirement analysis, \textbf{\texttt{D6.} \texttt{Represent all perspectives}} is not a priority. There could also be cases where the cause of subjectivity can be identified without knowing what exact type of subjectivity is being dealt with.

\section{Subjectivity Evaluation Practices}
\label{sec:current-eval-practices}
In Section~\ref{sec:desiderata}, we formulated the desiderata for a subjectivity-sensitive model in a top-down fashion. We now look bottom-up, exploring how these desiderata translate to concrete evaluation practices. We discuss existing metrics and analyze how current evaluation practices relate to our desiderata in both classification and free-text generation settings. 

\subsection{Existing Metrics}
\label{subsec:existing-metrics}
We first discuss metrics used to measure aspects of how well a model captures subjectivity in its predictions. We review related literature and identify the main metrics and evaluation practices.\footnote{Our aim is not to provide an exhaustive list of existing metrics, but to provide a snapshot of important metrics and evaluation methods to identify gaps the field can focus on.} 

\paragraph{Predicting the full human judgment distribution.} 
Determining whether subjectivity is accurately predicted depends on how the model \textit{expresses} it. When \textit{disagreement} is represented as \textit{soft} labels, we measure the distance between the soft labels and the softmax distribution of the model. Widely used metrics are (variations of) \textit{KL-divergence}, \textit{Jensen-Shannon Divergence}, \textit{Mean Absolute Error}, \textit{Wasserstein}, or \textit{Total Variation Distance} \citep{rizzi-etal-2024-soft}.  
To assess the spread of a single distribution, entropy $\mathcal{H}$ is used \citep{alies-etal-2025-measuring}.
Whether the model captures human disagreement, we can measure the correlation between the entropy of the soft label and the softmax distribution ($\mathcal{H}_{corr}$) \citep{Uma_Fornaciari_Hovy_Paun_Plank_Poesio_2020}. These metrics work for classification, but are not applicable when evaluating free-text generated responses. For generation, NLI models are used to determine whether the generated response covers the different necessary perspectives \citep{feng-etal-2024-modular, shetty-etal-2025-vital}. 

\paragraph{Single label scenarios.} 
Although not suitable to evaluate \textit{soft} labels, \textsc{accuracy} and \textsc{$F_1$-score} are still used in subjective scenarios, particularly when modeling an individual user (steering) or evaluating an aggregation of the subjective predictions to a majority label \citep{davani-etal-2022-dealing}. 
Additionally, looking at specific minorities may reveal performance insights on the least represented annotators \citep{van-der-meer-etal-2024-annotator}. \textsc{$F_1$-score} can also be used to determine whether a model is good at recognizing subjective input \citep{homayounirad-etal-2025-will}, or turned into soft versions \citep{kurniawan-hlv-empirical-2025}.  

\paragraph{Demographical bias.} 
Certain metrics measure the extent to which a model represents (or is biased toward) specific demographic groups.
This can be done by measuring the similarity in responses between a model and a specific country \citep{durmus2024towards} or demographic \citep{santurkar2023whose}, 
or using a linear mixed-effects model \citep{sun-etal-2025-sociodemographic}. \citet{jiang-etal-2025-language} propose \textsc{Value Inequity Index} for whether model responses are sensitive to an individual's demographic or not. 

\paragraph{Reliability.} Reliability metrics such as \textsc{correlation} and \textit{inter-annotator agreement} (IAA) are used for a variety of analyses. IAA is used mainly to measure the agreement between the predictions of the model and the human prediction \citep{masud-etal-2024-hate, hoeken-etal-2025-just}. Apart from $\mathcal{H}_{corr}$, 
most commonly, such metrics can be used to measure whether the behavior of an LLM globally corresponds to human ratings of items in the dataset \citep{sap-etal-2022-annotators, ryan-etal-2024-unintended, parappan-henao-2025-learning, chen-etal-2024-seeing}, to specific demographics \citep{giorgi-etal-2024-modeling, pihulski2025language} or for other fine-grained analyses \citep{guo2025counterfactual, luz-de-araujo-etal-2025-principled, plepi-etal-2022-unifying, giorgi-etal-2024-modeling, szekely2007distancecorrelation}.

\paragraph{Calibration.} A model is deemed well-calibrated when its confidence scores match its empirical performance. For subjective tasks, this means predicted probabilities match the human judgment distribution. While \textsc{distance metrics} already capture this, \citet{baan-etal-2022-stop} propose measuring on an instance-level the difference in entropy between the soft label and the model's predicted distribution and whether the class rankings match. 

\paragraph{Other.} The following metrics do not fall into the categories above or are too infrequent to be a separate category. \citet{santurkar2023whose} propose measuring \textit{consistency}; does the model's default output consistently reflect the same demographic across different topics? 
\textsc{Visualizing} important dynamics has also gained traction \citep{wright-etal-2024-llm, chen-etal-2024-seeing}.
Alternatively, LLM judges are used to evaluate whether a steered response is better than no steering, represented through the \textsc{Win Rate}: the proportion of samples for which the steered model is preferred \citep{chen2025pad, tan-lee-2025-unmasking, jang2024personalized}.
Finally, \citet{samuel-etal-2025-personagym} propose \texttt{PersonaScore}, a human-aligned score that evaluates the generated free-text response according to persona agent-specific tasks.

\begin{figure*}[h!]
    \centering
    \subfloat[The number of papers that cover each desiderata.]{%
        \includegraphics[width=0.45\textwidth]{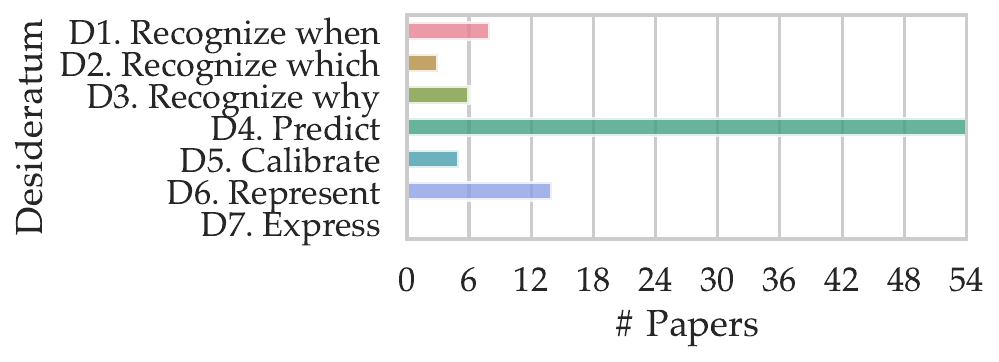}
        \label{fig:desideratum-papers}
    }
    \hfill
    \subfloat[The number of papers that cover each metric type.]{%
        \includegraphics[width=0.45\textwidth]{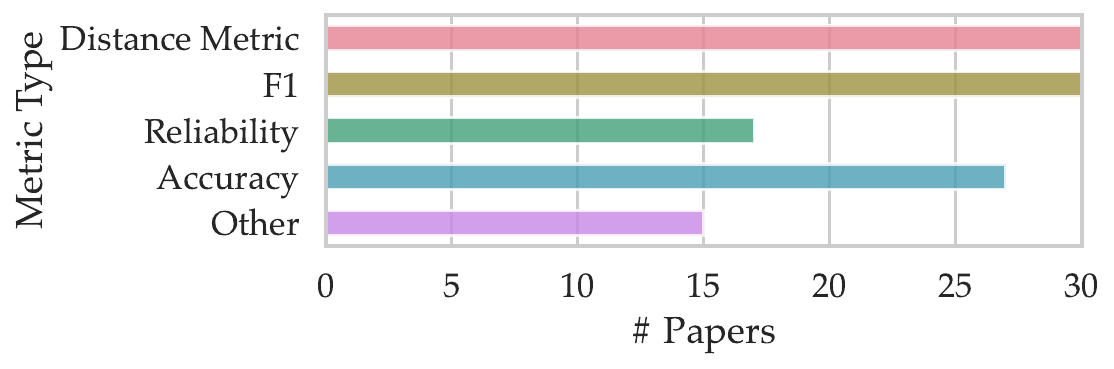}
        \label{fig:metric-type-papers}
    }
    \hfill
    \subfloat[The number of unique metric types used and desiderata investigated by the amount of papers.]{%
        \includegraphics[width=0.45\textwidth]{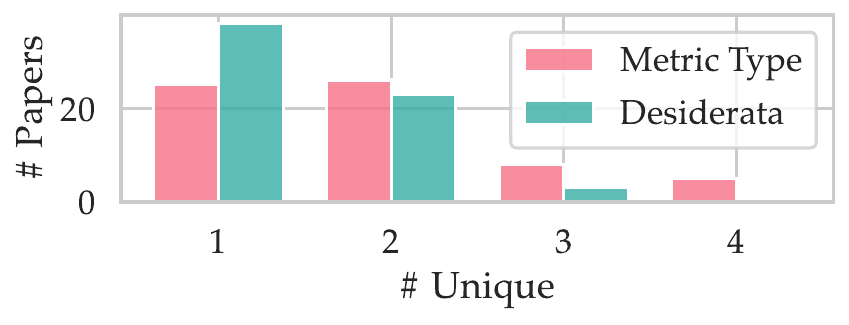}
        \label{fig:n-metric-type-papers}
    }
    \hfill
    \subfloat[Number of papers that address a specific evaluation objective.]{%
        \includegraphics[width=0.45\textwidth]{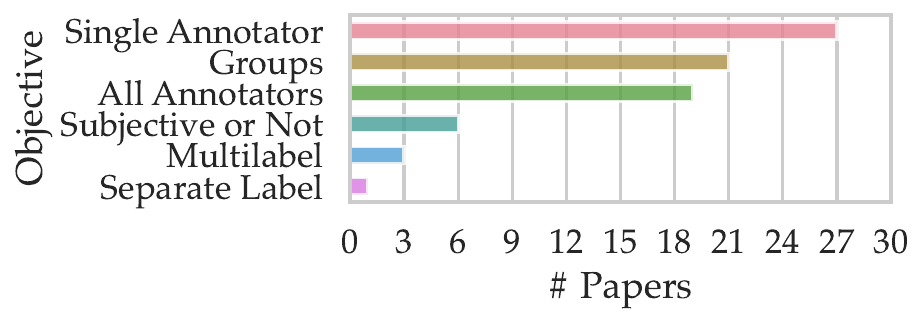}
        \label{fig:objective-papers}
    }
    \caption{Statistics from mapping subjective evaluation practices to our desiderata. We examine \npapersreviewed unique papers, however note that a paper can be counted under multiple metric types and desiderata.}
    \label{fig:paper-stats}
\end{figure*}

\subsection{Metrics and Desiderata in Papers}
\label{subsec:mapping-metrics-desiderata}

We analyze how the metrics are used in practice in relation to our desiderata. Based on the metrics we discussed, we distinguish the following categorization of metrics that are regularly used for evaluating aspects of subjectivity: \textsc{Distance Metric}, \textsc{$F_1$}, \textsc{Accuracy}, \textsc{Reliability}, and \textsc{Other}.   

We describe our literature search and annotation method in Appendix~\ref{app:literature-search-annotation}. We found \npapersreviewed~papers and annotated them, indicating the category of metrics used, the evaluation objective, and the covered desiderata.
Figure~\ref{fig:paper-stats} shows our mapping. These visualizations are meant to give a general idea of evaluation practices; the distributions presented in the figures are based on an indicative (not exhaustive) literature search.

\paragraph{\texttt{D1.}\texttt{Recognize when there is subjectivity}} is captured as a binary classification task (subjective or not) or as a continuous task (how subjective). In the binary case, \textsc{$F_1$} \citep{wan2023everyone, homayounirad-etal-2025-will} and \textsc{Accuracy} are used \citep{romberg-2022-perspective}. Depending on how the continuous label is represented, evaluation ranges from using a distance metric \citep{wan2023everyone} to $\mathcal{H}_{corr}$. 

\paragraph{\texttt{D2.}\texttt{Recognize which type of subjectivity}} is not explicitly evaluated in the papers we considered. Few studies compare performance by subjectivity type by either manually labeling each sample’s subjectivity type \citep{jiang-marneffe-2022-investigating, sandri-etal-2023-dont} or using datasets tailored to specific types of subjectivity \citep{feng-etal-2024-modular}.

\paragraph{\texttt{D3.}\texttt{Recognize why the input is subjective}} is evaluated in a single study, which tests whether a model can generate and recognize disambiguations in NLI \citep{liu-etal-2023-afraid}. For generation, \textsc{Edit-$F_1$} (an $F_1$-variant accounting for added and deleted unigrams between reference and generation) was used and \textsc{Accuracy} for the recognition of disambiguations.
Other papers conducted a qualitative analysis of keywords responsible for prediction using feature attribution \citep{wan2023everyone, muscato2025perspectives} or manually annotated the reasons for subjectivity and compared model performance accordingly \citep{sandri-etal-2023-dont, hoeken-etal-2025-just}.

\paragraph{\texttt{D4.}\texttt{Predict subjectivity accurately}} is evaluated through different metric categories, as discussed in Section~\ref{subsec:existing-metrics}. Depending on the evaluation objective, variants of \textsc{Accuracy}, \textsc{F1}, \textsc{Reliability}, \textsc{Distance Metric}, and \textsc{Other} (\textsc{Visualization} or \textsc{Win Rate}) are used. 

\paragraph{\texttt{D5.}\texttt{Calibrate the output with respect to frequency of opinions}} is mainly evaluated via \textsc{Distance Metrics}. Some papers evaluate calibration in terms of model confidence matching empirical performance \citep{sorensen-etal-2025-value, parappan-henao-2025-learning}. However, such evaluations are not specific to subjectivity. 

\paragraph{\texttt{D6.}\texttt{Represent all perspectives}} is evaluated from a variety of angles. 
\textsc{Distance Metrics} are used to compare differences in predictions between demographics or to measure similarity between a model's default output and specific demographics. \textsc{Reliability} metrics are used to investigate the correlation between the model's default output and a specific perspective and to measure whether the model can represent demographic groups well. \textsc{Accuracy} is used for per-country or demographic performance or \textsc{Value Inequity Index} to measure the extent to which model performance deviates across different demographics.

\paragraph{In general,} we find that the papers in our analysis focused primarily on \texttt{D4.}\texttt{Predict} and did not investigate \texttt{D7.}\texttt{Express} subjectivity at all (Figure~\ref{fig:desideratum-papers}). The most prevalent types of metric are \textsc{distance metric}, $F_1$, and \textsc{Accuracy} (Figure~\ref{fig:metric-type-papers}). Single annotators are most popular as the evaluation objective, followed by representing groups (Figure~\ref{fig:objective-papers}). Modeling the full human judgment distribution, not conditioned on a demographic, is also well-studied. Most papers consider one to two metric types or desiderata (Figure~\ref{fig:n-metric-type-papers}).

\section{Opportunities and Challenges}
\label{sec:opportunities-challenges}
We discuss the gaps (\includegraphics[height=1em]{icons/gap.png}) we identified that present opportunities (\includegraphics[height=1em]{icons/ideas.png}) for future work along with critical challenges, keeping our desiderata in mind. 
 
\paragraph{\includegraphics[height=1em]{icons/gap.png} The desideratum \texttt{D7.Express} is unaddressed.} 
None of the papers in our analysis study how to express subjectivity to the user without hampering their experience. Effective modeling of subjectivity is of limited value if the outcome is not adequately conveyed to the user.
We can borrow insights and metrics from other fields, e.g., Human-Computer Interaction (HCI) \citep{ehsan-2020-hcxai, benjamin-expl-2022}, Affective Computing \citep{kunc-ac-2024}, or Computer-Supported Collaborative Work \citep{lai-selective-explanations-2023}, to guide new expression evaluation strategies. \\ \textbf{\includegraphics[height=1.25em]{icons/ideas.png} \textit{Use insights and metrics from adjacent fields to evaluate \texttt{D7.}\texttt{Express}.}}

\paragraph{\includegraphics[height=1em]{icons/gap.png} The desiderata \texttt{D2.}\texttt{Recognize which} and \texttt{D3.}\texttt{Recognize why} are understudied.} 
Subjectivity type (\texttt{D2}) and reasons for subjectivity (\texttt{D3}) are addressed implicitly through performance-based analyses. For \texttt{D3}, only one paper explicitly evaluates disambiguation (in NLI). Some papers resort to qualitative feature-attribution analyses. Interpretability or explainability techniques can play a role in identifying key drivers in the input toward the prediction made. We propose combining these methods with manually annotated data to construct metrics that verify whether the model is making its predictions for the right reasons.

To study \texttt{D2}, we can start from the inherent uncertainty of the input. Traditionally, neural network uncertainty is decomposed into \textit{aleatoric} and \textit{epistemic} uncertainty \citep{hullermeier2021aleatoric}. Aleatoric uncertainty arises from the input itself, such as ambiguity or noise; measuring it can help identify ambiguous inputs. It remains challenging to measure the uncertainty surrounding the polyphony of an input and we invite future work to tackle this. \\
\textbf{\includegraphics[height=1.25em]{icons/ideas.png} \textit{Design metrics to address \texttt{D2.}\texttt{Recognize which} and \texttt{D3.}\texttt{Recognize why}, potentially through methodologies for uncertainty estimation and explainability.}}

\paragraph{\includegraphics[height=1em]{icons/gap.png} The entanglement of polyphony and ambiguity has not been at the forefront yet.} These two have been studied in isolation.  As illustrated in Section~\ref{sec:dimensions}, the two types of subjectivity can co-occur and understanding how a model handles this interplay can reveal broader insights into its sensitivity to subjective input. \\
\textbf{\includegraphics[height=1.25em]{icons/ideas.png} \textit{Study the entanglement of polyphony and ambiguity at a sample-level.}}

\paragraph{\includegraphics[height=1em]{icons/gap.png} A multi-faceted and deployment-driven investigation of model behavior is missing.} Considering \textit{all} desiderata is challenging due to limited evaluation practices and experimental constraints. 
Yet, these aspects should not be studied in isolation but their \textit{interplay} must be studied as well. We find that papers do not evaluate the consequences of earlier desiderata on the subsequent desiderata (e.g., the impact of the model performing badly on \texttt{D3.}\texttt{Recognize why} on \texttt{D4.}\texttt{Predict}), while this is crucial for ensuring reliable model deployment.  \textbf{\includegraphics[height=1.25em]{icons/ideas.png} \textit{Study the interplay of different desiderata.}}

Further, NLP researchers may not know how the model will express subjectivity, particularly in the case of free-text generation scenarios (e.g., overton vs steerable). Current research directions emphasize steering or personalizing a model toward a specific perspective, rather than investigating how the model fares in capturing a wider or even the (full) spectrum of perspectives. Thus, it is essential to evaluate all three operationalizations of \textit{pluralistic alignment} for informed deployment decisions. \textbf{\includegraphics[height=1.25em]{icons/ideas.png} \textit{Evaluate modes of expressing subjectivity.}}

\paragraph{\includegraphics[height=1em]{icons/gap.png} Evaluation of \texttt{D4.}\texttt{Predict} in free-text generation is limited.} Most metrics are directly applicable in classification scenarios, but require additional steps for free-text-generated responses. Future work can focus on extending metrics beyond NLI- and LLM-as-a-judge-based solutions. 
Using LLMs to evaluate has been found to be unreliable as it can come with biases, such as length, order, or self-preferences \citep{koo-etal-2024-benchmarking}, and can also amplify human-like biases that we would like to mitigate, such as gender or misinformation oversight bias \citep{chen-etal-2024-humans}.
\\ \textbf{\includegraphics[height=1.25em]{icons/ideas.png} \textit{Design metrics to evaluate how well free-text generation predicts subjectivity.}}. 

\paragraph{\includegraphics[height=1em]{icons/gap.png} Current datasets still fail to meet standards for subjective evaluation.} Datasets are increasingly being released with disaggregated labels and annotator backgrounds (e.g., \citet{sachdeva-etal-2022-measuring, aroyo-dices-2023}), including information such as annotator confidence ratings \citep{alacam-etal-2025-disentangling}. However, this might not be sufficient to capture the full spectrum of perspectives. Answering questions such as \textit{who} should annotate \textit{what} requires interdisciplinary efforts, grounding annotation strategies in fields such as intersectional feminism \cite{10.1145/3531146.3533207}, moral philosophy \citep{Graham2013}, or social science \citep{zhao2021sample}. Future developments include metrics that can measure the diversity of the annotator distribution for a given sample or datasets with annotations that enable high-quality subjective representations in the model and thorough evaluation approaches. \\
\noindent \textbf{\includegraphics[height=1.25em]{icons/ideas.png} \textit{Develop interdisciplinary approaches to improve datasets and annotation strategies.}}

\section{Conclusion}
This paper analyzes the current evaluation paradigms of subjectivity-sensitive models.
We begin by understanding subjectivity and how it is represented in NLP data and models, which leads us to define \textbf{desiderata} for a subjectivity-sensitive model. We take a top-down approach to design \textit{seven} desiderata, driven by our ideals for a model to (1) represent all, based on (2) the correct source of subjectivity to (3) generalize to the real world. 
We analyze \npapersreviewed~papers to map evaluation metrics to our desiderata. This allows us to identify gaps that can be fruitful directions for future work. Among these gaps, we find that existing work has overlooked nuances in the types of subjectivity: polyphony versus ambiguity, a distinction we argue is crucial for effective evaluation. Additionally, current work does not focus on the desideratum: \texttt{Express}, nor the interplay between different desiderata. 

\section*{Limitations}
\paragraph{Scope of Evaluation Directions} 
This paper aims to demonstrate that the evaluation of subjectivity-sensitive studies remains an open challenge.
Our primary objective was to identify key gaps in the evaluation of subjectivity-sensitive models and outline potential directions for future research.
Although we discuss potential strategies for evaluating such models throughout the paper, the implementation of concrete evaluation methodologies is left for future work. However, inspiration from the concrete metrics in Section~\ref{subsec:existing-metrics} can be taken as a starting point.

\paragraph{Literature Survey Methodology} 
To support the discussion of the desiderata introduced in this work, we survey state-of-the-art work in Section~\ref{subsec:mapping-metrics-desiderata}. We emphasize that the goal of this survey is not to provide an exhaustive overview of the field, as done by \citet{frenda2025perspectivist, xu2026beyond}, but rather to obtain a broad characterization of current evaluation practices. The selection of metrics (i.e., annotation labels) for the surveyed papers was performed in a top-down manner to align with our proposed desiderata. While this choice enables consistent alignment within our paper, it may also have caused certain patterns to be overlooked; for more comprehensive surveys, we refer the reader to \citet{frenda2025perspectivist, xu2026beyond}. Finally, each paper was annotated by a single author, which may introduce some degree of annotation subjectivity. To mitigate this issue, the authors who annotated discussed and calibrated the annotation process after division of papers on a subset of papers (6 samples between one set of two authors and 8 samples between another set of two others, amounting to roughly 10\% of the total amount of papers).

\section*{Acknowledgments}
We thank the anonymous reviewers for their comments that helped improve this paper. All remaining errors are our own.
This research was (partially) funded by the Hybrid Intelligence Center, a 10-year programme, and AlgoSoc, a collaborative 10-year research program on public values in the algorithmic society, both funded by the Dutch Ministry of Education, Culture and Science under the Gravitation programme (project numbers 024.005.017 and 024.004.022) through the Netherlands Organisation for Scientific Research. Any opinions, findings, and
conclusions or recommendations expressed in this material are those of the author(s) and do not necessarily reflect the views of OCW or those of the AlgoSoc consortium as a whole. The figures and affiliation emojis have been designed using resources from flaticon.com; the chatbot for \textit{Express} and person behind laptop are by Flat Icons, lightbulb + hand by Dewi Sari, and the rest of the emojis
by freepik.

\bibliography{anthology_0, anthology_1, custom}

@article{
zhang2025personalization,
title={Personalization of Large Language Models: A Survey},
author={Zhehao Zhang and Ryan A. Rossi and Branislav Kveton and Yijia Shao and Diyi Yang and Hamed Zamani and Franck Dernoncourt and Joe Barrow and Tong Yu and Sungchul Kim and Ruiyi Zhang and Jiuxiang Gu and Tyler Derr and Hongjie Chen and Junda Wu and Xiang Chen and Zichao Wang and Subrata Mitra and Nedim Lipka and Nesreen K. Ahmed and Yu Wang},
journal={Transactions on Machine Learning Research},
issn={2835-8856},
year={2025},
url={https://openreview.net/forum?id=tf6A9EYMo6},
note={Survey Certification}
}

@inproceedings{sorensen2024position,
author = {Sorensen, Taylor and Moore, Jared and Fisher, Jillian and Gordon, Mitchell and Mireshghallah, Niloofar and Rytting, Christopher Michael and Ye, Andre and Jiang, Liwei and Lu, Ximing and Dziri, Nouha and Althoff, Tim and Choi, Yejin},
title = {Position: a roadmap to pluralistic alignment},
year = {2024},
publisher = {JMLR.org},
booktitle = {Proceedings of the 41st International Conference on Machine Learning},
articleno = {1882},
numpages = {23},
location = {Vienna, Austria},
series = {ICML'24}
}

@inproceedings{
durmus2024towards,
title={Towards Measuring the Representation of Subjective Global Opinions in Language Models},
author={Esin Durmus and Karina Nguyen and Thomas Liao and Nicholas Schiefer and Amanda Askell and Anton Bakhtin and Carol Chen and Zac Hatfield-Dodds and Danny Hernandez and Nicholas Joseph and Liane Lovitt and Sam McCandlish and Orowa Sikder and Alex Tamkin and Janel Thamkul and Jared Kaplan and Jack Clark and Deep Ganguli},
booktitle={First Conference on Language Modeling},
year={2024},
url={https://openreview.net/forum?id=zl16jLb91v}
}

@inproceedings{sun-etal-2025-sociodemographic,
    title = "Sociodemographic Prompting is Not Yet an Effective Approach for Simulating Subjective Judgments with {LLM}s",
    author = "Sun, Huaman  and
      Pei, Jiaxin  and
      Choi, Minje  and
      Jurgens, David",
    editor = "Chiruzzo, Luis  and
      Ritter, Alan  and
      Wang, Lu",
    booktitle = "Proceedings of the 2025 Conference of the Nations of the Americas Chapter of the Association for Computational Linguistics: Human Language Technologies (Volume 2: Short Papers)",
    month = apr,
    year = "2025",
    address = "Albuquerque, New Mexico",
    publisher = "Association for Computational Linguistics",
    url = "https://aclanthology.org/2025.naacl-short.71/",
    doi = "10.18653/v1/2025.naacl-short.71",
    pages = "845--854",
    ISBN = "979-8-89176-190-2"
}

@article{aroyo2015truth,
  title={Truth is a lie: Crowd truth and the seven myths of human annotation},
  author={Aroyo, Lora and Welty, Chris},
  journal={AI Magazine},
  volume={36},
  number={1},
  pages={15--24},
  year={2015}
}

@book{vardomskaya2018sources,
  title={Sources of subjectivity},
  author={Vardomskaya, Tamara Nikolaevna},
  year={2018},
  publisher={The University of Chicago}
}

@inproceedings{
jang2024personalized,
title={Personalized Soups: Personalized Large Language Model Alignment via Post-hoc Parameter Merging},
author={Joel Jang and Seungone Kim and Bill Yuchen Lin and Yizhong Wang and Jack Hessel and Luke Zettlemoyer and Hannaneh Hajishirzi and Yejin Choi and Prithviraj Ammanabrolu},
booktitle={Adaptive Foundation Models: Evolving AI for Personalized and Efficient Learning},
year={2024},
url={https://openreview.net/forum?id=EMrnoPRvxe}
}

@inproceedings{orlikowski-etal-2025-beyond,
    title = "Beyond Demographics: Fine-tuning Large Language Models to Predict Individuals' Subjective Text Perceptions",
    author = {Orlikowski, Matthias  and
      Pei, Jiaxin  and
      R{\"o}ttger, Paul  and
      Cimiano, Philipp  and
      Jurgens, David  and
      Hovy, Dirk},
    editor = "Che, Wanxiang  and
      Nabende, Joyce  and
      Shutova, Ekaterina  and
      Pilehvar, Mohammad Taher",
    booktitle = "Proceedings of the 63rd Annual Meeting of the Association for Computational Linguistics (Volume 1: Long Papers)",
    month = jul,
    year = "2025",
    address = "Vienna, Austria",
    publisher = "Association for Computational Linguistics",
    url = "https://aclanthology.org/2025.acl-long.104/",
    doi = "10.18653/v1/2025.acl-long.104",
    pages = "2092--2111",
    ISBN = "979-8-89176-251-0"
}

@inproceedings{jurylearning-gordon-2022,
author = {Gordon, Mitchell L. and Lam, Michelle S. and Park, Joon Sung and Patel, Kayur and Hancock, Jeff and Hashimoto, Tatsunori and Bernstein, Michael S.},
title = {Jury Learning: Integrating Dissenting Voices into Machine Learning Models},
year = {2022},
isbn = {9781450391573},
publisher = {Association for Computing Machinery},
address = {New York, NY, USA},
url = {https://doi.org/10.1145/3491102.3502004},
doi = {10.1145/3491102.3502004},
booktitle = {Proceedings of the 2022 CHI Conference on Human Factors in Computing Systems},
articleno = {115},
numpages = {19},
location = {New Orleans, LA, USA},
series = {CHI '22}
}

@inproceedings{crowdworksheets-diaz-2022,
author = {D\'{\i}az, Mark and Kivlichan, Ian and Rosen, Rachel and Baker, Dylan and Amironesei, Razvan and Prabhakaran, Vinodkumar and Denton, Remi},
title = {CrowdWorkSheets: Accounting for Individual and Collective Identities Underlying Crowdsourced Dataset Annotation},
year = {2022},
isbn = {9781450393522},
publisher = {Association for Computing Machinery},
address = {New York, NY, USA},
url = {https://doi.org/10.1145/3531146.3534647},
doi = {10.1145/3531146.3534647},
booktitle = {Proceedings of the 2022 ACM Conference on Fairness, Accountability, and Transparency},
pages = {2342–2351},
numpages = {10},
location = {Seoul, Republic of Korea},
series = {FAccT '22}
}

@article{toxicity-goyal-2022,
author = {Goyal, Nitesh and Kivlichan, Ian D. and Rosen, Rachel and Vasserman, Lucy},
title = {Is Your Toxicity My Toxicity? Exploring the Impact of Rater Identity on Toxicity Annotation},
year = {2022},
issue_date = {November 2022},
publisher = {Association for Computing Machinery},
address = {New York, NY, USA},
volume = {6},
number = {CSCW2},
url = {https://doi.org/10.1145/3555088},
doi = {10.1145/3555088},
journal = {Proc. ACM Hum.-Comput. Interact.},
month = nov,
articleno = {363},
numpages = {28}
}

@inproceedings{rlhfpaper,
 author = {Ouyang, Long and Wu, Jeffrey and Jiang, Xu and Almeida, Diogo and Wainwright, Carroll and Mishkin, Pamela and Zhang, Chong and Agarwal, Sandhini and Slama, Katarina and Ray, Alex and Schulman, John and Hilton, Jacob and Kelton, Fraser and Miller, Luke and Simens, Maddie and Askell, Amanda and Welinder, Peter and Christiano, Paul F and Leike, Jan and Lowe, Ryan},
 booktitle = {Advances in Neural Information Processing Systems},
 editor = {S. Koyejo and S. Mohamed and A. Agarwal and D. Belgrave and K. Cho and A. Oh},
 pages = {27730--27744},
 publisher = {Curran Associates, Inc.},
 title = {Training language models to follow instructions with human feedback},
 url = {https://proceedings.neurips.cc/paper_files/paper/2022/file/b1efde53be364a73914f58805a001731-Paper-Conference.pdf},
 volume = {35},
 year = {2022}
}

@inproceedings{dpopaper,
 author = {Rafailov, Rafael and Sharma, Archit and Mitchell, Eric and Manning, Christopher D and Ermon, Stefano and Finn, Chelsea},
 booktitle = {Advances in Neural Information Processing Systems},
 editor = {A. Oh and T. Naumann and A. Globerson and K. Saenko and M. Hardt and S. Levine},
 pages = {53728--53741},
 publisher = {Curran Associates, Inc.},
 title = {Direct Preference Optimization: Your Language Model is Secretly a Reward Model},
 url = {https://proceedings.neurips.cc/paper_files/paper/2023/file/a85b405ed65c6477a4fe8302b5e06ce7-Paper-Conference.pdf},
 volume = {36},
 year = {2023}
}

@inproceedings{rodrigues2018deep,
author = {Rodrigues, Filipe and Pereira, Francisco C.},
title = {Deep learning from crowds},
year = {2018},
isbn = {978-1-57735-800-8},
publisher = {AAAI Press},
booktitle = {Proceedings of the Thirty-Second AAAI Conference on Artificial Intelligence and Thirtieth Innovative Applications of Artificial Intelligence Conference and Eighth AAAI Symposium on Educational Advances in Artificial Intelligence},
articleno = {197},
numpages = {8},
location = {New Orleans, Louisiana, USA},
series = {AAAI'18/IAAI'18/EAAI'18}
}

@inproceedings{
khurana2024crowdcalibrator,
title={Crowd-Calibrator: Can Annotator Disagreement Inform Calibration in Subjective Tasks?},
author={Urja Khurana and Eric Nalisnick and Antske Fokkens and Swabha Swayamdipta},
booktitle={First Conference on Language Modeling},
year={2024},
url={https://openreview.net/forum?id=VWWzO3ewMS}
}

@inproceedings{santurkar2023whose,
author = {Santurkar, Shibani and Durmus, Esin and Ladhak, Faisal and Lee, Cinoo and Liang, Percy and Hashimoto, Tatsunori},
title = {Whose opinions do language models reflect?},
year = {2023},
publisher = {JMLR.org},
booktitle = {Proceedings of the 40th International Conference on Machine Learning},
articleno = {1244},
numpages = {34},
location = {Honolulu, Hawaii, USA},
series = {ICML'23}
}

@article{frenda2025perspectivist,
  title={Perspectivist approaches to natural language processing: a survey},
  author={Frenda, Simona and Abercrombie, Gavin and Basile, Valerio and Pedrani, Alessandro and Panizzon, Raffaella and Cignarella, Alessandra Teresa and Marco, Cristina and Bernardi, Davide},
  journal={Language Resources and Evaluation},
  volume={59},
  number={2},
  pages={1719--1746},
  year={2025},
  publisher={Springer}
}

@inproceedings{pandita2025forest,
  author={Deepak Pandita and Flip Korn and Chris Welty and Christopher M. Homan},
  title={Forest vs Tree: The (N, K) Trade-off in Reproducible ML Evaluation},
  year={2026},
  cdate={1767225600000},
  pages={24736-24744},
  url={https://doi.org/10.1609/aaai.v40i29.39659},
  booktitle={AAAI},
}

@article{Uma_Fornaciari_Hovy_Paun_Plank_Poesio_2020, title={A Case for Soft Loss Functions}, volume={8}, url={https://ojs.aaai.org/index.php/HCOMP/article/view/7478}, DOI={10.1609/hcomp.v8i1.7478}, abstractNote={&lt;p class=&quot;abstract&quot;&gt;Recently, Peterson et al. provided evidence of the benefits of using probabilistic soft labels generated from crowd annotations for training a computer vision model, showing that using such labels maximizes performance of the models over unseen data. In this paper, we generalize these results by showing that training with soft labels is an effective method for using crowd annotations in several other ai tasks besides the one studied by Peterson &lt;em&gt;et al.&lt;/em&gt;, and also when their performance is compared with that of state-of-the-art methods for learning from crowdsourced data.&lt;/p&gt;}, number={1}, journal={Proceedings of the AAAI Conference on Human Computation and Crowdsourcing}, author={Uma, Alexandra and Fornaciari, Tommaso and Hovy, Dirk and Paun, Silviu and Plank, Barbara and Poesio, Massimo}, year={2020}, month={Oct.}, pages={173-177} }

@inproceedings{lourie2021scruples,
  title={Scruples: A corpus of community ethical judgments on 32,000 real-life anecdotes},
  author={Lourie, Nicholas and Le Bras, Ronan and Choi, Yejin},
  booktitle={Proceedings of the AAAI Conference on Artificial Intelligence},
  volume={35},
  number={15},
  pages={13470--13479},
  year={2021}
}

@inproceedings{homayounirad-etal-2025-will,
    title = "Will Annotators Disagree? Identifying Subjectivity in Value-Laden Arguments",
    author = "Homayounirad, Amir  and
      Liscio, Enrico  and
      Wang, Tong  and
      Jonker, Catholijn M  and
      Siebert, Luciano Cavalcante",
    editor = "Christodoulopoulos, Christos  and
      Chakraborty, Tanmoy  and
      Rose, Carolyn  and
      Peng, Violet",
    booktitle = "Findings of the Association for Computational Linguistics: EMNLP 2025",
    month = nov,
    year = "2025",
    address = "Suzhou, China",
    publisher = "Association for Computational Linguistics",
    url = "https://aclanthology.org/2025.findings-emnlp.824/",
    doi = "10.18653/v1/2025.findings-emnlp.824",
    pages = "15237--15252",
    ISBN = "979-8-89176-335-7"
}

@article{hullermeier2021aleatoric,
  title={Aleatoric and epistemic uncertainty in machine learning: An introduction to concepts and methods},
  author={H{\"u}llermeier, Eyke and Waegeman, Willem},
  journal={Machine learning},
  volume={110},
  number={3},
  pages={457--506},
  year={2021},
  publisher={Springer}
}

@inproceedings{
guo2025counterfactual,
title={Counterfactual Reasoning for Steerable Pluralistic Value Alignment of Large Language Models},
author={Hanze Guo and Jing Yao and Xiao Zhou and Xiaoyuan Yi and Xing Xie},
booktitle={The Thirty-ninth Annual Conference on Neural Information Processing Systems},
year={2025},
url={https://openreview.net/forum?id=86b23oNkg9}
}

@article{uma2021learning,
  title={Learning from disagreement: A survey},
  author={Uma, Alexandra N and Fornaciari, Tommaso and Hovy, Dirk and Paun, Silviu and Plank, Barbara and Poesio, Massimo},
  journal={Journal of Artificial Intelligence Research},
  volume={72},
  doi={https://doi.org/10.1613/jair.1.12752},
  pages={1385--1470},
  year={2021}
}

@article{cabitza2023toward, title={Toward a Perspectivist Turn in Ground Truthing for Predictive Computing}, volume={37}, url={https://ojs.aaai.org/index.php/AAAI/article/view/25840}, DOI={10.1609/aaai.v37i6.25840}, abstractNote={Most current Artificial Intelligence applications are based on supervised Machine Learning (ML), which ultimately grounds on data annotated by small teams of experts or large ensemble of volunteers. The annotation process is often performed in terms of a majority vote, however this has been proved to be often problematic by recent evaluation studies.
In this article, we describe and advocate for a different paradigm, which we call perspectivism: this counters the removal of disagreement and, consequently, the assumption of correctness of traditionally aggregated gold-standard datasets, and proposes the adoption of methods that preserve divergence of opinions and integrate multiple perspectives in the ground truthing process of ML development. Drawing on previous works which inspired it, mainly from the crowdsourcing and multi-rater labeling settings, we survey the state-of-the-art and describe the potential of our proposal for not only the more subjective tasks (e.g. those related to human language) but also those tasks commonly understood as objective (e.g. medical decision making). We present the main benefits of adopting a perspectivist stance in ML, as well as possible disadvantages, and various ways in which such a stance can be implemented in practice. Finally, we share a set of recommendations and outline a research agenda to advance the perspectivist stance in ML.}, number={6}, journal={Proceedings of the AAAI Conference on Artificial Intelligence}, author={Cabitza, Federico and Campagner, Andrea and Basile, Valerio}, year={2023}, month={Jun.}, pages={6860-6868} }

@inproceedings{jiang-etal-2025-language,
    title = "Can Language Models Reason about Individualistic Human Values and Preferences?",
    author = "Jiang, Liwei  and
      Sorensen, Taylor  and
      Levine, Sydney  and
      Choi, Yejin",
    editor = "Che, Wanxiang  and
      Nabende, Joyce  and
      Shutova, Ekaterina  and
      Pilehvar, Mohammad Taher",
    booktitle = "Proceedings of the 63rd Annual Meeting of the Association for Computational Linguistics (Volume 1: Long Papers)",
    month = jul,
    year = "2025",
    address = "Vienna, Austria",
    publisher = "Association for Computational Linguistics",
    url = "https://aclanthology.org/2025.acl-long.336/",
    doi = "10.18653/v1/2025.acl-long.336",
    pages = "6757--6794",
    ISBN = "979-8-89176-251-0"
}

@inproceedings{shetty-etal-2025-vital,
    title = "{VITAL}: A New Dataset for Benchmarking Pluralistic Alignment in Healthcare",
    author = "Shetty, Anudeex  and
      Beheshti, Amin  and
      Dras, Mark  and
      Naseem, Usman",
    editor = "Che, Wanxiang  and
      Nabende, Joyce  and
      Shutova, Ekaterina  and
      Pilehvar, Mohammad Taher",
    booktitle = "Proceedings of the 63rd Annual Meeting of the Association for Computational Linguistics (Volume 1: Long Papers)",
    month = jul,
    year = "2025",
    address = "Vienna, Austria",
    publisher = "Association for Computational Linguistics",
    url = "https://aclanthology.org/2025.acl-long.1119/",
    doi = "10.18653/v1/2025.acl-long.1119",
    pages = "22954--22974",
    ISBN = "979-8-89176-251-0"
}

@inproceedings{
chen2025pad,
title={{PAD}: Personalized Alignment at Decoding-time},
author={Ruizhe Chen and Xiaotian Zhang and Meng Luo and Wenhao Chai and Zuozhu Liu},
booktitle={The Thirteenth International Conference on Learning Representations},
year={2025},
url={https://openreview.net/forum?id=e7AUJpP8bV}
}

@inproceedings{tan-lee-2025-unmasking,
    title = "Unmasking Implicit Bias: Evaluating Persona-Prompted {LLM} Responses in Power-Disparate Social Scenarios",
    author = "Tan, Bryan Chen Zhengyu  and
      Lee, Roy Ka-Wei",
    editor = "Chiruzzo, Luis  and
      Ritter, Alan  and
      Wang, Lu",
    booktitle = "Proceedings of the 2025 Conference of the Nations of the Americas Chapter of the Association for Computational Linguistics: Human Language Technologies (Volume 1: Long Papers)",
    month = apr,
    year = "2025",
    address = "Albuquerque, New Mexico",
    publisher = "Association for Computational Linguistics",
    url = "https://aclanthology.org/2025.naacl-long.50/",
    doi = "10.18653/v1/2025.naacl-long.50",
    pages = "1075--1108",
    ISBN = "979-8-89176-189-6"
}

@inproceedings{samuel-etal-2025-personagym,
    title = "{P}ersona{G}ym: Evaluating Persona Agents and {LLM}s",
    author = "Samuel, Vinay  and
      Zou, Henry Peng  and
      Zhou, Yue  and
      Chaudhari, Shreyas  and
      Kalyan, Ashwin  and
      Rajpurohit, Tanmay  and
      Deshpande, Ameet  and
      Narasimhan, Karthik R  and
      Murahari, Vishvak",
    editor = "Christodoulopoulos, Christos  and
      Chakraborty, Tanmoy  and
      Rose, Carolyn  and
      Peng, Violet",
    booktitle = "Findings of the Association for Computational Linguistics: EMNLP 2025",
    month = nov,
    year = "2025",
    address = "Suzhou, China",
    publisher = "Association for Computational Linguistics",
    url = "https://aclanthology.org/2025.findings-emnlp.368/",
    doi = "10.18653/v1/2025.findings-emnlp.368",
    pages = "6999--7022",
    ISBN = "979-8-89176-335-7"
}

@inproceedings{dumitrache2018crowdtruth,
  title={CrowdTruth 2.0: Quality metrics for crowdsourcing with disagreement},
  author={Dumitrache, Anca and Inel, Oana and Aroyo, Lora and Timmermans, Benjamin and Welty, Chris},
  booktitle={1st Workshop on Subjectivity, Ambiguity and Disagreement in Crowdsourcing, and Short Paper 1st Workshop on Disentangling the Relation Between Crowdsourcing and Bias Management, SAD+ CrowdBias 2018},
  pages={11--18},
  year={2018},
  organization={CEUR-WS}
}

@article{szekely2007distancecorrelation,
 ISSN = {00905364},
 URL = {http://www.jstor.org/stable/25464608},
 author = {Gábor J. Székely and Maria L. Rizzo and Nail K. Bakirov},
 journal = {The Annals of Statistics},
 number = {6},
 pages = {2769--2794},
 publisher = {Institute of Mathematical Statistics},
 title = {Measuring and Testing Dependence by Correlation of Distances},
 urldate = {2025-12-10},
 volume = {35},
 year = {2007}
}

@inproceedings{
poddar2024personalizing,
title={Personalizing Reinforcement Learning from Human Feedback with Variational Preference Learning},
author={Sriyash Poddar and Yanming Wan and Hamish Ivison and Abhishek Gupta and Natasha Jaques},
booktitle={The Thirty-eighth Annual Conference on Neural Information Processing Systems},
year={2024},
url={https://openreview.net/forum?id=gRG6SzbW9p}
}

@inproceedings{meister-etal-2025-benchmarking,
    title = "Benchmarking Distributional Alignment of Large Language Models",
    author = "Meister, Nicole  and
      Guestrin, Carlos  and
      Hashimoto, Tatsunori",
    editor = "Chiruzzo, Luis  and
      Ritter, Alan  and
      Wang, Lu",
    booktitle = "Proceedings of the 2025 Conference of the Nations of the Americas Chapter of the Association for Computational Linguistics: Human Language Technologies (Volume 1: Long Papers)",
    month = apr,
    year = "2025",
    address = "Albuquerque, New Mexico",
    publisher = "Association for Computational Linguistics",
    url = "https://aclanthology.org/2025.naacl-long.2/",
    doi = "10.18653/v1/2025.naacl-long.2",
    pages = "24--49",
    ISBN = "979-8-89176-189-6"
}

@inproceedings{
zhao2024group,
title={Group Preference Optimization: Few-Shot Alignment of Large Language Models},
author={Siyan Zhao and John Dang and Aditya Grover},
booktitle={The Twelfth International Conference on Learning Representations},
year={2024},
url={https://openreview.net/forum?id=DpFeMH4l8Q}
}

@article{
casper2023open,
title={Open Problems and Fundamental Limitations of Reinforcement Learning from Human Feedback},
author={Stephen Casper and Xander Davies and Claudia Shi and Thomas Krendl Gilbert and J{\'e}r{\'e}my Scheurer and Javier Rando and Rachel Freedman and Tomek Korbak and David Lindner and Pedro Freire and Tony Tong Wang and Samuel Marks and Charbel-Raphael Segerie and Micah Carroll and Andi Peng and Phillip J.K. Christoffersen and Mehul Damani and Stewart Slocum and Usman Anwar and Anand Siththaranjan and Max Nadeau and Eric J Michaud and Jacob Pfau and Dmitrii Krasheninnikov and Xin Chen and Lauro Langosco and Peter Hase and Erdem Biyik and Anca Dragan and David Krueger and Dorsa Sadigh and Dylan Hadfield-Menell},
journal={Transactions on Machine Learning Research},
issn={2835-8856},
year={2023},
url={https://openreview.net/forum?id=bx24KpJ4Eb},
note={Survey Certification, Featured Certification}
}

@INPROCEEDINGS{pihulski2025language,
  author={Pihulski, Dzmitry and Koco{\'n}, Jan},
  booktitle={2025 IEEE International Conference on Data Mining Workshops (ICDMW)}, 
  title={Language, Culture, and Ideology: Personalizing Offensiveness Detection in Political Tweets with Reasoning LLMs}, 
  year={2025},
  volume={},
  number={},
  pages={2146-2155},
  doi={10.1109/ICDMW69685.2025.00261}}

@inproceedings{parappan-henao-2025-learning,
    title = "Learning Subjective Label Distributions via Sociocultural Descriptors",
    author = "Parappan, Mohammed Fayiz  and
      Henao, Ricardo",
    editor = "Christodoulopoulos, Christos  and
      Chakraborty, Tanmoy  and
      Rose, Carolyn  and
      Peng, Violet",
    booktitle = "Proceedings of the 2025 Conference on Empirical Methods in Natural Language Processing",
    month = nov,
    year = "2025",
    address = "Suzhou, China",
    publisher = "Association for Computational Linguistics",
    url = "https://aclanthology.org/2025.emnlp-main.1026/",
    doi = "10.18653/v1/2025.emnlp-main.1026",
    pages = "20322--20338",
    ISBN = "979-8-89176-332-6"
}

@inproceedings{luz-de-araujo-etal-2025-principled,
    title = "Principled Personas: Defining and Measuring the Intended Effects of Persona Prompting on Task Performance",
    author = {Luz de Araujo, Pedro Henrique  and
      R{\"o}ttger, Paul  and
      Hovy, Dirk  and
      Roth, Benjamin},
    editor = "Christodoulopoulos, Christos  and
      Chakraborty, Tanmoy  and
      Rose, Carolyn  and
      Peng, Violet",
    booktitle = "Proceedings of the 2025 Conference on Empirical Methods in Natural Language Processing",
    month = nov,
    year = "2025",
    address = "Suzhou, China",
    publisher = "Association for Computational Linguistics",
    url = "https://aclanthology.org/2025.emnlp-main.1364/",
    doi = "10.18653/v1/2025.emnlp-main.1364",
    pages = "26857--26886",
    ISBN = "979-8-89176-332-6"
}

@inproceedings{ehsan-2020-hcxai,
author = {Ehsan, Upol and Riedl, Mark O.},
title = {Human-Centered Explainable AI: Towards a Reflective Sociotechnical Approach},
year = {2020},
isbn = {978-3-030-60116-4},
publisher = {Springer-Verlag},
address = {Berlin, Heidelberg},
url = {https://doi.org/10.1007/978-3-030-60117-1_33},
doi = {10.1007/978-3-030-60117-1_33},
booktitle = {HCI International 2020 - Late Breaking Papers: Multimodality and Intelligence: 22nd HCI International Conference, HCII 2020, Copenhagen, Denmark, July 19–24, 2020, Proceedings},
pages = {449–466},
numpages = {18},
location = {Copenhagen, Denmark}
}

@article{xu2026beyond,
  title={Beyond Consensus: Perspectivist Modeling and Evaluation of Annotator Disagreement in NLP},
  author={Xu, Yinuo and Jurgens, David},
  journal={arXiv preprint arXiv:2601.09065},
  year={2026}
}

@inproceedings{ivey-etal-2025-nutmeg,
    title = "{NUTMEG}: Separating Signal From Noise in Annotator Disagreement",
    author = "Ivey, Jonathan  and
      Gauch, Susan  and
      Jurgens, David",
    editor = "Christodoulopoulos, Christos  and
      Chakraborty, Tanmoy  and
      Rose, Carolyn  and
      Peng, Violet",
    booktitle = "Proceedings of the 2025 Conference on Empirical Methods in Natural Language Processing",
    month = nov,
    year = "2025",
    address = "Suzhou, China",
    publisher = "Association for Computational Linguistics",
    url = "https://aclanthology.org/2025.emnlp-main.144/",
    doi = "10.18653/v1/2025.emnlp-main.144",
    pages = "2874--2887",
    ISBN = "979-8-89176-332-6"
}

@inproceedings{abercrombie-etal-2025-consistency,
    title = "Consistency is Key: Disentangling Label Variation in Natural Language Processing with Intra-Annotator Agreement",
    author = "Abercrombie, Gavin  and
      Dinkar, Tanvi  and
      Cercas Curry, Amanda  and
      Rieser, Verena  and
      Hovy, Dirk",
    editor = "Abercrombie, Gavin  and
      Basile, Valerio  and
      Frenda, Simona  and
      Tonelli, Sara  and
      Dudy, Shiran",
    booktitle = "Proceedings of the The 4th Workshop on Perspectivist Approaches to NLP",
    month = nov,
    year = "2025",
    address = "Suzhou, China",
    publisher = "Association for Computational Linguistics",
    url = "https://aclanthology.org/2025.nlperspectives-1.6/",
    doi = "10.18653/v1/2025.nlperspectives-1.6",
    pages = "63--74",
    ISBN = "979-8-89176-350-0"
}

@inproceedings{ignatev-etal-2025-hypernetworks,
    title = "Hypernetworks for Perspectivist Adaptation",
    author = "Ignatev, Daniil  and
      Paperno, Denis  and
      Poesio, Massimo",
    editor = "Abercrombie, Gavin  and
      Basile, Valerio  and
      Frenda, Simona  and
      Tonelli, Sara  and
      Dudy, Shiran",
    booktitle = "Proceedings of the The 4th Workshop on Perspectivist Approaches to NLP",
    month = nov,
    year = "2025",
    address = "Suzhou, China",
    publisher = "Association for Computational Linguistics",
    url = "https://aclanthology.org/2025.nlperspectives-1.10/",
    doi = "10.18653/v1/2025.nlperspectives-1.10",
    pages = "111--122",
    ISBN = "979-8-89176-350-0"
}

@inproceedings{disagreement-deconvolution-gordon-2021,
author = {Gordon, Mitchell L. and Zhou, Kaitlyn and Patel, Kayur and Hashimoto, Tatsunori and Bernstein, Michael S.},
title = {The Disagreement Deconvolution: Bringing Machine Learning Performance Metrics In Line With Reality},
year = {2021},
isbn = {9781450380966},
publisher = {Association for Computing Machinery},
address = {New York, NY, USA},
url = {https://doi.org/10.1145/3411764.3445423},
doi = {10.1145/3411764.3445423},
booktitle = {Proceedings of the 2021 CHI Conference on Human Factors in Computing Systems},
articleno = {388},
numpages = {14},
location = {Yokohama, Japan},
series = {CHI '21}
}

@inproceedings{alies-etal-2025-measuring,
    title = "Measuring Label Ambiguity in Subjective Tasks using Predictive Uncertainty Estimation",
    author = "Alies, Richard  and
      Merdjanovska, Elena  and
      Akbik, Alan",
    editor = "Peng, Siyao  and
      Rehbein, Ines",
    booktitle = "Proceedings of the 19th Linguistic Annotation Workshop (LAW-XIX-2025)",
    month = jul,
    year = "2025",
    address = "Vienna, Austria",
    publisher = "Association for Computational Linguistics",
    url = "https://aclanthology.org/2025.law-1.2/",
    doi = "10.18653/v1/2025.law-1.2",
    pages = "21--34",
    ISBN = "979-8-89176-262-6"
}

@inproceedings{alacam-etal-2025-disentangling,
    title = "Disentangling Subjectivity and Uncertainty for Hate Speech Annotation and Modeling using Gaze",
    author = {Alacam, {\"O}zge  and
      Hoeken, Sanne  and
      S{\"a}uberli, Andreas  and
      Gr{\"o}ner, Hannes  and
      Frassinelli, Diego  and
      Zarrie{\ss}, Sina  and
      Plank, Barbara},
    editor = "Christodoulopoulos, Christos  and
      Chakraborty, Tanmoy  and
      Rose, Carolyn  and
      Peng, Violet",
    booktitle = "Proceedings of the 2025 Conference on Empirical Methods in Natural Language Processing",
    month = nov,
    year = "2025",
    address = "Suzhou, China",
    publisher = "Association for Computational Linguistics",
    url = "https://aclanthology.org/2025.emnlp-main.1460/",
    doi = "10.18653/v1/2025.emnlp-main.1460",
    pages = "28707--28724",
    ISBN = "979-8-89176-332-6"
}

@article{kurniawan-hlv-empirical-2025,
    author = {Kurniawan, Kemal and Mistica, Meladel and Baldwin, Timothy and Lau, Jey Han},
    title = {Training and Evaluating with Human Label Variation: An Empirical Study},
    journal = {Computational Linguistics},
    volume = {52},
    number = {1},
    pages = {85-111},
    year = {2026},
    month = {03},
    issn = {0891-2017},
    doi = {10.1162/COLI.a.578},
    url = {https://doi.org/10.1162/COLI.a.578},
    eprint = {https://direct.mit.edu/coli/article-pdf/52/1/85/2561976/coli.a.578.pdf},
}

@inproceedings{hoeken-etal-2025-just,
    title = "Not Just Who or What: Modeling the Interaction of Linguistic and Annotator Variation in Hateful Word Interpretation",
    author = {Hoeken, Sanne  and
      Alacam, {\"O}zge  and
      Nguyen, Dong  and
      Poesio, Massimo  and
      Zarrie{\ss}, Sina},
    editor = "Evang, Kilian  and
      Kallmeyer, Laura  and
      Pogodalla, Sylvain",
    booktitle = "Proceedings of the 16th International Conference on Computational Semantics",
    month = sep,
    year = "2025",
    address = {D{\"u}sseldorf, Germany},
    publisher = "Association for Computational Linguistics",
    url = "https://aclanthology.org/2025.iwcs-main.6/",
    pages = "63--77",
    ISBN = "979-8-89176-316-6"
}

@inproceedings{wan2023everyone,
author = {Wan, Ruyuan and Kim, Jaehyung and Kang, Dongyeop},
title = {Everyone's voice matters: quantifying annotation disagreement using demographic information},
year = {2023},
isbn = {978-1-57735-880-0},
publisher = {AAAI Press},
url = {https://doi.org/10.1609/aaai.v37i12.26698},
doi = {10.1609/aaai.v37i12.26698},
booktitle = {Proceedings of the Thirty-Seventh AAAI Conference on Artificial Intelligence and Thirty-Fifth Conference on Innovative Applications of Artificial Intelligence and Thirteenth Symposium on Educational Advances in Artificial Intelligence},
articleno = {1629},
numpages = {8},
series = {AAAI'23/IAAI'23/EAAI'23}
}

@inproceedings{muscato2025perspectives,
  title     = {Perspectives in Play: A Multi-Perspective Approach for More Inclusive NLP Systems},
  author    = {Muscato, Benedetta and Passaro, Lucia and Gezici, Gizem and Giannotti, Fosca},
  booktitle = {Proceedings of the Thirty-Fourth International Joint Conference on
               Artificial Intelligence, {IJCAI-25}},
  publisher = {International Joint Conferences on Artificial Intelligence Organization},
  editor    = {James Kwok},
  pages     = {9827--9835},
  year      = {2025},
  month     = {8},
  note      = {AI and Social Good},
  doi       = {10.24963/ijcai.2025/1092},
  url       = {https://doi.org/10.24963/ijcai.2025/1092},
}

@incollection{Graham2013,
    title = {{Moral Foundations Theory: The Pragmatic Validity of Moral Pluralism}},
    year = {2013},
    booktitle = {Advances in Experimental Social Psychology},
    author = {Graham, Jesse and Haidt, Jonathan and Koleva, Sena and Motyl, Matt and Iyer, Ravi and Wojcik, Sean P. and Ditto, Peter H.},
    pages = {55--130},
    volume = {47},
    publisher = {Elsevier},
    ignore_address = {Amsterdam, the Netherlands},
    ignore_doi = {10.1016/B978-0-12-407236-7.00002-4},
}

@inproceedings{10.1145/3531146.3533207,
author = {Klumbyt\`{e}, Goda and Draude, Claude and Taylor, Alex S.},
title = {Critical Tools for Machine Learning: Working with Intersectional Critical Concepts in Machine Learning Systems Design},
year = {2022},
isbn = {9781450393522},
publisher = {Association for Computing Machinery},
address = {New York, NY, USA},
url = {https://doi.org/10.1145/3531146.3533207},
doi = {10.1145/3531146.3533207},
booktitle = {Proceedings of the 2022 ACM Conference on Fairness, Accountability, and Transparency},
pages = {1528–1541},
numpages = {14},
location = {Seoul, Republic of Korea},
series = {FAccT '22}
}

@inproceedings{sorensen-etal-2025-value,
    title = "Value Profiles for Encoding Human Variation",
    author = "Sorensen, Taylor  and
      Mishra, Pushkar  and
      Patel, Roma  and
      Tessler, Michael Henry  and
      Bakker, Michiel A.  and
      Evans, Georgina  and
      Gabriel, Iason  and
      Goodman, Noah  and
      Rieser, Verena",
    editor = "Christodoulopoulos, Christos  and
      Chakraborty, Tanmoy  and
      Rose, Carolyn  and
      Peng, Violet",
    booktitle = "Proceedings of the 2025 Conference on Empirical Methods in Natural Language Processing",
    month = nov,
    year = "2025",
    address = "Suzhou, China",
    publisher = "Association for Computational Linguistics",
    url = "https://aclanthology.org/2025.emnlp-main.106/",
    doi = "10.18653/v1/2025.emnlp-main.106",
    pages = "2047--2095",
    ISBN = "979-8-89176-332-6"
}

@INPROCEEDINGS{kunc-ac-2024,
  author={Kunc, Dominika and Komoszyńska, Joanna and Saganowski, Stanisław and Kazienko, Przemysław and Quigley, Karen S. and Barrett, Lisa Feldman},
  booktitle={2024 12th International Conference on Affective Computing and Intelligent Interaction Workshops and Demos (ACIIW)}, 
  title={Embracing Subjectivity in Affective Research: Naturalistic and Controlled Settings}, 
  year={2024},
  volume={},
  number={},
  pages={176-180},
  doi={10.1109/ACIIW63320.2024.00035}}

@inproceedings{wang-etal-2025-perspective,
    title = "Perspective Transition of Large Language Models for Solving Subjective Tasks",
    author = "Wang, Xiaolong  and
      Zhang, Yuanchi  and
      Wang, Ziyue  and
      Xu, Yuzhuang  and
      Luo, Fuwen  and
      Wang, Yile  and
      Li, Peng  and
      Liu, Yang",
    editor = "Che, Wanxiang  and
      Nabende, Joyce  and
      Shutova, Ekaterina  and
      Pilehvar, Mohammad Taher",
    booktitle = "Findings of the Association for Computational Linguistics: ACL 2025",
    month = jul,
    year = "2025",
    address = "Vienna, Austria",
    publisher = "Association for Computational Linguistics",
    url = "https://aclanthology.org/2025.findings-acl.502/",
    doi = "10.18653/v1/2025.findings-acl.502",
    pages = "9686--9704",
    ISBN = "979-8-89176-256-5"
}

@article{benjamin-expl-2022,
author = {Benjamin, Jesse Josua and Kinkeldey, Christoph and M\"{u}ller-Birn, Claudia and Korjakow, Tim and Herbst, Eva-Maria},
title = {Explanation Strategies as an Empirical-Analytical Lens for Socio-Technical Contextualization of Machine Learning Interpretability},
year = {2022},
issue_date = {January 2022},
publisher = {Association for Computing Machinery},
address = {New York, NY, USA},
volume = {6},
number = {GROUP},
url = {https://doi.org/10.1145/3492858},
doi = {10.1145/3492858},
journal = {Proc. ACM Hum.-Comput. Interact.},
month = jan,
articleno = {39},
numpages = {25}
}

@article{lai-selective-explanations-2023,
author = {Lai, Vivian and Zhang, Yiming and Chen, Chacha and Liao, Q. Vera and Tan, Chenhao},
title = {Selective Explanations: Leveraging Human Input to Align Explainable AI},
year = {2023},
issue_date = {October 2023},
publisher = {Association for Computing Machinery},
address = {New York, NY, USA},
volume = {7},
number = {CSCW2},
url = {https://doi.org/10.1145/3610206},
doi = {10.1145/3610206},
journal = {Proc. ACM Hum.-Comput. Interact.},
month = oct,
articleno = {357},
numpages = {35}
}

@inproceedings{aroyo-dices-2023,
 author = {Aroyo, Lora and Taylor, Alex and D\'{\i}az, Mark and Homan, Christopher and Parrish, Alicia and Serapio-Garc\'{\i}a, Gregory and Prabhakaran, Vinodkumar and Wang, Ding},
 booktitle = {Advances in Neural Information Processing Systems},
 editor = {A. Oh and T. Naumann and A. Globerson and K. Saenko and M. Hardt and S. Levine},
 pages = {53330--53342},
 publisher = {Curran Associates, Inc.},
 title = {DICES Dataset: Diversity in Conversational AI Evaluation for Safety},
 url = {https://proceedings.neurips.cc/paper_files/paper/2023/file/a74b697bce4cac6c91896372abaa8863-Paper-Datasets_and_Benchmarks.pdf},
 volume = {36},
 year = {2023}
}

@article{zhao2021sample,
  title={Sample representation in the social sciences},
  author={Zhao, Kino},
  journal={Synthese},
  volume={198},
  number={10},
  pages={9097--9115},
  year={2021},
  publisher={Springer}
}

@inproceedings{
poole-dayan2025benchmarking,
title={Benchmarking Overton Pluralism in {LLM}s},
author={Elinor Poole-Dayan and Jiayi Wu and Jiaxin Pei and Michiel A. Bakker},
booktitle={NeurIPS 2025 Workshop on Evaluating the Evolving LLM Lifecycle: Benchmarks, Emergent Abilities, and Scaling},
year={2025},
url={https://openreview.net/forum?id=Mq3QUs7FAp}
}

@inproceedings{zhou-etal-2025-modeling,
    title = "Modeling Subjectivity in Cognitive Appraisal with Language Models",
    author = "Zhou, Yuxiang  and
      Xu, Hainiu  and
      Ong, Desmond C.  and
      Liakata, Maria  and
      Slovak, Petr  and
      He, Yulan",
    editor = "Christodoulopoulos, Christos  and
      Chakraborty, Tanmoy  and
      Rose, Carolyn  and
      Peng, Violet",
    booktitle = "Findings of the Association for Computational Linguistics: EMNLP 2025",
    month = nov,
    year = "2025",
    address = "Suzhou, China",
    publisher = "Association for Computational Linguistics",
    url = "https://aclanthology.org/2025.findings-emnlp.744/",
    doi = "10.18653/v1/2025.findings-emnlp.744",
    pages = "13811--13833",
    ISBN = "979-8-89176-335-7"
}

@inproceedings{gupta2024personabias,
    title = {Bias {R}uns {D}eep: Implicit Reasoning Biases in Persona-Assigned {LLM}s},
    author = {Gupta, Shashank and Shrivastava, Vaishnavi and Deshpande, Ameet and Kalyan, Ashwin and Clark, Peter and Sabharwal, Ashish and Khot, Tushar},
    booktitle = {The Twelfth International Conference on Learning Representations},
    year = {2024},
    url={https://openreview.net/forum?id=kGteeZ18Ir}
}

@article{wang2025large,
  title={Large language models that replace human participants can harmfully misportray and flatten identity groups},
  author={Wang, Angelina and Morgenstern, Jamie and Dickerson, John P},
  journal={Nature Machine Intelligence},
  volume={7},
  number={3},
  pages={400--411},
  year={2025},
  publisher={Nature Publishing Group UK London}
}

@inproceedings{
    li2025prefpalette,
    title={PrefPalette: Personalized Preference Modeling with Latent Attributes},
    author={Shuyue Stella Li and Melanie Sclar and Hunter Lang and Ansong Ni and Jacqueline He and Puxin Xu and Andrew Cohen and Chan Young Park and Yulia Tsvetkov and Asli Celikyilmaz},
    booktitle={Second Conference on Language Modeling},
    year={2025},
    url={https://openreview.net/forum?id=p4ujQsKmPV}
}

@inproceedings{lo-etal-2025-perseval,
    title = "{PERSEVAL}: A Framework for Perspectivist Classification Evaluation",
    author = "Lo, Soda Marem  and
      Casola, Silvia  and
      Sezerer, Erhan  and
      Basile, Valerio  and
      Sansonetti, Franco  and
      Uva, Antonio  and
      Bernardi, Davide",
    editor = "Christodoulopoulos, Christos  and
      Chakraborty, Tanmoy  and
      Rose, Carolyn  and
      Peng, Violet",
    booktitle = "Proceedings of the 2025 Conference on Empirical Methods in Natural Language Processing",
    month = nov,
    year = "2025",
    address = "Suzhou, China",
    publisher = "Association for Computational Linguistics",
    url = "https://aclanthology.org/2025.emnlp-main.1137/",
    doi = "10.18653/v1/2025.emnlp-main.1137",
    pages = "22334--22359",
    ISBN = "979-8-89176-332-6"
}

@inproceedings{miehling-etal-2025-evaluating,
    title = "Evaluating the Prompt Steerability of Large Language Models",
    author = "Miehling, Erik  and
      Desmond, Michael  and
      Natesan Ramamurthy, Karthikeyan  and
      Daly, Elizabeth M.  and
      Varshney, Kush R.  and
      Farchi, Eitan  and
      Dognin, Pierre  and
      Rios, Jesus  and
      Bouneffouf, Djallel  and
      Liu, Miao  and
      Sattigeri, Prasanna",
    editor = "Chiruzzo, Luis  and
      Ritter, Alan  and
      Wang, Lu",
    booktitle = "Proceedings of the 2025 Conference of the Nations of the Americas Chapter of the Association for Computational Linguistics: Human Language Technologies (Volume 1: Long Papers)",
    month = apr,
    year = "2025",
    address = "Albuquerque, New Mexico",
    publisher = "Association for Computational Linguistics",
    url = "https://aclanthology.org/2025.naacl-long.400/",
    doi = "10.18653/v1/2025.naacl-long.400",
    pages = "7874--7900",
    ISBN = "979-8-89176-189-6"
}

@inproceedings{lutz-etal-2025-prompt,
    title = "The Prompt Makes the Person(a): A Systematic Evaluation of Sociodemographic Persona Prompting for Large Language Models",
    author = "Lutz, Marlene  and
      Sen, Indira  and
      Ahnert, Georg  and
      Rogers, Elisa  and
      Strohmaier, Markus",
    editor = "Christodoulopoulos, Christos  and
      Chakraborty, Tanmoy  and
      Rose, Carolyn  and
      Peng, Violet",
    booktitle = "Findings of the Association for Computational Linguistics: EMNLP 2025",
    month = nov,
    year = "2025",
    address = "Suzhou, China",
    publisher = "Association for Computational Linguistics",
    url = "https://aclanthology.org/2025.findings-emnlp.1261/",
    doi = "10.18653/v1/2025.findings-emnlp.1261",
    pages = "23212--23237",
    ISBN = "979-8-89176-335-7"
}

@inproceedings{alam2025clef,
  title={The clef-2025 checkthat! lab: Subjectivity, fact-checking, claim normalization, and retrieval},
  author={Alam, Firoj and Stru{\ss}, Julia Maria and Chakraborty, Tanmoy and Dietze, Stefan and Hafid, Salim and Korre, Katerina and Muti, Arianna and Nakov, Preslav and Ruggeri, Federico and Schellhammer, Sebastian and others},
  booktitle={European Conference on Information Retrieval},
  pages={467--478},
  year={2025},
  organization={Springer}
}

@inproceedings{sarumi-etal-2025-impact,
    title = "The Impact of Annotator Personas on {LLM} Behavior Across the Perspectivism Spectrum",
    author = {Sarumi, Olufunke O.  and
      Welch, Charles  and
      Braun, Daniel  and
      Schl{\"o}tterer, J{\"o}rg},
    editor = "Abbas, Mourad  and
      Yousef, Tariq  and
      Galke, Lukas",
    booktitle = "Proceedings of the 8th International Conference on Natural Language and Speech Processing (ICNLSP-2025)",
    month = aug,
    year = "2025",
    address = "Southern Denmark University, Odense, Denmark",
    publisher = "Association for Computational Linguistics",
    url = "https://aclanthology.org/2025.icnlsp-1.14/",
    pages = "121--136"
}

@inproceedings{
zhang2025diverging,
title={Diverging Preferences: When do Annotators Disagree and do Models Know?},
author={Michael JQ Zhang and Zhilin Wang and Jena D. Hwang and Yi Dong and Olivier Delalleau and Yejin Choi and Eunsol Choi and Xiang Ren and Valentina Pyatkin},
booktitle={Forty-second International Conference on Machine Learning},
year={2025},
url={https://openreview.net/forum?id=qWgAAVhoXb}
}

@inproceedings{kocon2021learning,
  title={Learning personal human biases and representations for subjective tasks in natural language processing},
  author={Koco{\'n}, Jan and Gruza, Marcin and Bielaniewicz, Julita and Grimling, Damian and Kanclerz, Kamil and Mi{\l}kowski, Piotr and Kazienko, Przemys{\l}aw},
  booktitle={2021 IEEE international conference on data mining (ICDM)},
  pages={1168--1173},
  year={2021},
  organization={IEEE}
}

@inproceedings{pachinger-etal-2025-disaggregated,
    title = "A Disaggregated Dataset on {E}nglish Offensiveness Containing Spans",
    author = "Pachinger, Pia  and
      Goldzycher, Janis  and
      Planitzer, Anna M.  and
      Neidhardt, Julia  and
      Hanbury, Allan",
    editor = "Abercrombie, Gavin  and
      Basile, Valerio  and
      Frenda, Simona  and
      Tonelli, Sara  and
      Dudy, Shiran",
    booktitle = "Proceedings of the The 4th Workshop on Perspectivist Approaches to NLP",
    month = nov,
    year = "2025",
    address = "Suzhou, China",
    publisher = "Association for Computational Linguistics",
    url = "https://aclanthology.org/2025.nlperspectives-1.1/",
    doi = "10.18653/v1/2025.nlperspectives-1.1",
    pages = "1--14",
    ISBN = "979-8-89176-350-0"
}

\appendix

\section{Literature Review and Annotation}
\label{app:literature-search-annotation}
In Table~\ref{tab:annotations} we give an overview of the annotations for the papers we considered. 
We searched for papers on subjectivity in NLP through snowballing from key papers and search terms such as ``subjectivity nlp/llms'', ``disagreement nlp/llms'', ``perspectivism nlp'', ``persona prompting''. We found \npapersreviewed~papers and annotated them as follows: we indicated the category of metrics used (\textsc{distance metric}, \textsc{$F_1$}-variant, \textsc{reliability} metric, \textsc{accuracy}-variant, or \textsc{other}), the evaluation objective (single annotator/individual, all annotators/population, specific group (e.g., demographic based), subjective as a separate label, multi-label, or subjective or not) and the covered desiderata.

\paragraph{Inclusion and Exclusion Criteria.} We primarily focus on papers explicitly focusing on subjectivity, disagreement, or perspectivism in the field of NLP. For completeness, we also include papers focusing on persona prompting but leave out papers on \textit{personalization}. We do not limit our timeframe but prioritize recent papers. While we aimed to include only peer-reviewed work, we do also include preprints that give novel insights.

\paragraph{Annotation Procedure.} Each paper was annotated by one of the authors. To ensure alignment between the authors of the paper who annotated, we re-calibrated our decisions with each other on a handful of samples (6 samples between one of the two authors, and $8$ between another set of two others, roughly $10\%$) to ensure as much consistency as possible and align our understanding of the annotation dimensions. For annotating the desiderata that the paper addresses in its evaluation paradigm, we only consider what the paper explicitly says they evaluate for, not what they implicitly do without their intention (e.g., using distance metrics for whether subjectivity is done correctly but implicitly also checking calibration). 

\newcommand{\cc}{\ding{51}}
\newcommand{\xx}{\ding{55}}
\newcommand{\smc}{\texttt{D4}}
\newcommand{\unbiased}{\texttt{D6}}
\newcommand{\type}{\texttt{D2}}
\newcommand{\source}{\texttt{D3}}
\newcommand{\calibration}{\texttt{D5}}
\newcommand{\sm}{\texttt{D1}}
\newcommand{\single}{Individual}
\newcommand{\all}{All}
\newcommand{\group}{Groups}
\newcommand{\son}{Subjective or Not}
\newcommand{\separate}{Separate Label}
\newcommand{\multilabel}{Multilabel}

\begin{table*}[]
    \centering
    \scalebox{0.61}{
    \begin{tabular}{lrcccccr}
    \toprule
     & \textbf{\texttt{Desiderata}} & \textbf{\texttt{Reliability}} & \textbf{\texttt{Distance Metric}} & \textbf{F$_1$} & \textbf{\texttt{Accuracy}} & \textbf{\texttt{Other}} & \textbf{\texttt{Evaluation Objective}}  \\
    \midrule
    \citet{orlikowski-etal-2025-beyond} & \smc & \xx & \cc & \cc & \xx & \xx & \single \\
    \citet{van-der-meer-etal-2024-annotator} &  \smc & \xx & \cc & \cc & \xx & \xx & \all, \single \\
    \citet{wright-etal-2024-llm} & \smc, \unbiased & \xx & \cc & \xx & \xx & \textsc{Visualization} & \group \\
    \citet{feng-etal-2024-modular} & \type, \smc & \xx & \cc &	\cc & \cc & \textsc{Win Rate}, \textsc{NLI} & \group, \all \\
    \citet{wan2023everyone} & \sm, \source & \xx & \cc & \cc & \xx & \xx & \son \\ 
    \citet{poole-dayan2025benchmarking} & \smc & \xx & \cc & \xx & \cc & \xx & \all \\
    \citet{sorensen-etal-2025-value} & \smc & \xx & \cc & \xx & \cc & \xx & \all, \single \\ 
    \citet{jiang-marneffe-2022-investigating} & \sm, \type, \smc & \xx & \xx & \cc & \cc & \xx & \multilabel, \separate \\ 
    \citet{durmus2024towards} & \smc, \unbiased & \xx & \cc & \xx & \xx & \xx & \group \\
    \citet{sun-etal-2025-sociodemographic} & \smc, \unbiased & \xx & \xx & \xx & \xx & \textsc{Demographic} & \group \\ 
    \citet{giorgi-etal-2024-modeling} & \smc, \unbiased & \cc & \xx & \xx & \xx & \xx & \group \\
    \citet{zhou-etal-2025-modeling} & \smc, \calibration & \xx & \cc & \xx & \xx & \xx & \all \\
    \citet{dong-etal-2024-llm} & \smc & \xx & \xx & \xx & \cc & \xx & \single \\ 
    \citet{orlikowski-etal-2023-ecological} & \smc & \xx & \xx & \cc & \xx & \xx & \group \\ 
    \citet{santurkar2023whose} & \smc, \unbiased & \xx & \cc & \xx & \xx & \textsc{Consistency} & \group \\ 
    \citet{gupta2024personabias} & \unbiased & \xx & \xx & \xx & \cc & \xx & \group \\
    \citet{wang2025large} & \source, \smc & \xx & \cc & \xx & \xx & \xx & \group \\
    \citet{li2025prefpalette} & \smc & \xx & \xx & \xx & \cc & \xx & \group \\
    \citet{homayounirad-etal-2025-will} & \smc & \cc & \xx & \cc & \xx & \xx & \single, \son \\
    \citet{lo-etal-2025-perseval} & \smc, \unbiased & \xx & \xx & \cc & \xx & \xx & \all, \group, \single \\ 
    \citet{chen2025pad} & \smc & \xx & \xx & \xx & \xx & \textsc{Win Rate} & \single \\
    \citet{jurylearning-gordon-2022} & \smc & \xx & \cc & \xx & \cc & \xx & \single \\
    \citet{tan-lee-2025-unmasking} & \smc & \xx & \cc & \xx & \xx & \textsc{Win Rate} & \group \\ 
    \citet{samuel-etal-2025-personagym} & \smc & \xx & \xx & \xx & \xx & \textsc{PersonaScore} & \single \\ 
    \citet{miehling-etal-2025-evaluating} & \smc & \xx & \cc & \xx & \xx & \xx & \single \\ 
    \citet{chen-etal-2024-seeing} & \smc, \calibration & \cc & \cc & \cc & \xx & \textsc{Visualization} & \all \\
    \citet{lutz-etal-2025-prompt} & \smc & \xx & \cc & \xx & \xx & \xx & \group \\
    \citet{alam2025clef} & \sm & \xx & \xx & \cc & \xx & \xx & \son \\ 
    \citet{muscato2025perspectives} & \source, \smc & \xx & \cc & \cc & \cc & \textsc{Explainability} & \all \\
    \citet{sarumi-etal-2025-impact} & \smc & \xx & \cc & \cc & \xx & \xx & \single \\ 
    \citet{lee-etal-2023-large} & \smc, \calibration & \xx & \cc & \xx & \cc & \xx & \all \\ 
    \citet{liu-etal-2023-afraid} & \source, \smc & \xx & \xx & \cc & \cc & \xx & \multilabel \\
    \citet{plepi-etal-2022-unifying} & \smc, \unbiased & \cc & \xx & \cc & \cc & \xx & \single \\ 
    \citet{zhang2025diverging} & \sm, \smc, \unbiased & \xx & \cc & \xx & \cc & \xx & \single \\ 
    \citet{guo2025counterfactual} & \smc, \calibration & \cc & \cc & \xx & \xx & \xx & \multilabel \\ 
    \citet{pihulski2025language} & \smc & \cc & \xx & \xx & \xx & \xx & \single \\ 
    \citet{meister-etal-2025-benchmarking} & \smc, \calibration & \xx & \cc & \xx & \xx & \xx & \all \\ 
    \citet{zhao2024group} & \smc & \xx & \cc & \xx & \cc & \xx & \group, \single \\ 
    \citet{luz-de-araujo-etal-2025-principled} & \smc & \cc & \xx & \xx & \xx & \cc & \single \\ 
    \citet{parappan-henao-2025-learning} & \smc & \cc & \xx & \cc & \xx & \cc & \all, \group \\
    \citet{ryan-etal-2024-unintended} & \smc, \unbiased & \cc & \cc & \cc & \cc & \xx & \group \\ 
    \citet{fornaciari-etal-2021-beyond} & \smc & \xx & \xx & \cc & \cc & \xx & \all \\ 
    \citet{fleisig-etal-2023-majority} & \smc & \xx & \cc & \xx & \xx & \xx & \group, \single \\ 
    \citet{deng-etal-2023-annotate} & \smc & \xx & \xx & \cc & \cc & \xx & \single \\ 
    \citet{ignatev-etal-2025-hypernetworks} & \smc & \cc & \xx & \cc & \xx & \xx & \single \\ 
    \citet{weerasooriya-etal-2023-disagreement} & \smc & \xx & \cc & \cc & \cc & \xx & \all \\ 
    \citet{kocon2021learning} & \smc & \xx & \xx & \cc & \cc & \xx & \single \\ 
    \citet{sandri-etal-2023-dont} & \sm, \type, \source & \xx & \xx & \cc & \xx & \xx & \son \\ 
    \citet{mokhberian-etal-2024-capturing} & \smc & \cc & \xx & \cc & \xx & \textsc{Visualization} & \single \\ 
    \citet{alies-etal-2025-measuring} & \sm, \smc & \cc & \cc & \cc & \xx & \xx & \son \\ 
    \citet{pachinger-etal-2025-disaggregated} & \smc & \xx & \xx & \cc & \xx & \xx & \all \\  
    \citet{jang2024personalized} & \smc & \xx & \xx & \xx & \xx & \textsc{Win Rate} & \single \\ 
    \citet{hoeken-etal-2025-just} & \source, \smc & \cc & \xx & \xx & \cc & \textsc{Feature Association} & \single \\ 
    \citet{kurniawan-hlv-empirical-2025} & \smc & \cc & \cc & \cc & \cc & \xx & \all \\ 
    \citet{romberg-2022-perspective} & \sm & \xx & \xx & \cc & \cc & \xx & \son \\ 
    \citet{lee-etal-2024-exploring-cross} & \smc, \unbiased & \xx & \xx & \xx & \cc & \xx & \group \\ 
    \citet{masud-etal-2024-hate} & \smc, \unbiased & \cc & \xx & \cc & \xx & \xx & \group, \single \\ 
    \citet{davani-etal-2022-dealing} & \smc & \xx & \xx & \cc & \xx & \xx & \all, \single \\ 
    \citet{hwang-etal-2023-aligning} & \smc & \xx & \cc & \xx & \cc & \xx & \single \\ 
    \citet{jiang-etal-2025-language} & \smc, \unbiased & \xx & \xx & \xx & \cc & \xx & \single \\
    \citet{sap-etal-2022-annotators} & \unbiased & \cc & \xx & \xx & \xx & \xx & \group \\
    \citet{Uma_Fornaciari_Hovy_Paun_Plank_Poesio_2020} & \smc & \cc & \cc & \xx & \cc & \xx & \all \\
    \citet{uma-etal-2021-semeval} & \smc & \xx & \cc & \cc & \xx & \xx & \all \\ 
    \citet{shetty-etal-2025-vital} & \smc & \xx & \cc & \xx & \cc & \textsc{Win Rate}, \textsc{NLI} & \all, \group\\
    \bottomrule
    \end{tabular}}
\caption{Overview of papers considered and their annotations. \textbf{\texttt{Desiderata}}: the desiderata explicitly addressed by the paper, with the following options; \texttt{D1. Recognize when}, \texttt{D2. Recognize which type}, \texttt{D3. Recognize why }, \texttt{D4. Predict}, \texttt{D5. Calibrate}, \texttt{D6. Represent}, and \texttt{D7. Express}. \textbf{\texttt{Reliability}}: whether the paper uses a reliability metric. \textbf{\texttt{Distance Metric}}: whether the paper uses a distance metric. \textbf{\texttt{F$_1$}}: whether the paper uses $F_1$. \textbf{\texttt{Accuracy}}: whether the paper uses accuracy. \textbf{\texttt{Other}}: whether the paper uses any other metric, if so, which one? \textbf{\texttt{Evaluation objective}}: who is the paper evaluating for, with the following options: individual annotator, full human judgment distribution, a specific demographic or group, predicting whether an input is subjective or not, subjective is a separate label to be predicted, and multi-label predictions.}
\label{tab:annotations}
\end{table*}    

\end{document}